\documentclass[]{academic_template}

\usepackage[toc,page,header]{appendix}

\usepackage{amssymb}
\usepackage{amsmath}

\usepackage{makecell}
\usepackage{algorithm}      % 算法浮动环境
\usepackage{algpseudocode}  % 算法伪代码语法

\usepackage{graphicx}

\usepackage{listings}
\usepackage{url}
\usepackage{booktabs}
\usepackage{wrapfig}

\usepackage{colortbl}
\usepackage{xcolor}

\usepackage{caption}
\usepackage{subcaption}

\newcommand{\rmnum}[1]{\romannumeral #1}
\newcommand{\Rmnum}[1]{\expandafter\@slowromancap\romannumeral #1@}

\usepackage{bbding}
\usepackage{amssymb}
\usepackage{pifont}% http://ctan.org/pkg/pifont
\newcommand{\cmark}{\ding{51}}%
\newcommand{\xmark}{\ding{55}}%

\usepackage{multirow}
\usepackage[table,xcdraw]{xcolor}
\useunder{\uline}{\ul}{}

\setlogoheight{14mm}   % logo size
\setlogospacing{5mm}
\setsjtublue % 设置logo为上交蓝，默认为上交红

\settitlerulethickness{3pt}  % title line size

\abstractboxon %显示边框
\setabstractframecolor{gray} %灰色边框
\setabstractbgcolor{gray!10} %浅灰色背景
\setlogotolineshift{5mm} % logo与标题行的距离

\settoprulethickness{2.5pt}  
\setbottomrulethickness{1.5pt}

\title{Exploiting Unlabeled Data with Multiple Expert Teachers for Open Vocabulary Aerial Object Detection and Its Orientation Adaptation}

\author[1]{Yan Li}
\author[2, *]{Weiwei Guo}
\author[1, \dagger]{Xue Yang}
\author[1]{Ning Liao}
\author[1]{Shaofeng Zhang}
\author[3]{Yi Yu}
\author[1, *]{Wenxian Yu}
\author[1]{Junchi Yan}

\affiliation[1]{Shanghai Jiao Tong Univerisity}
\affiliation[2]{Tongji University}
\affiliation[3]{Southeast University}

\contribution[*]{Corresponding Author}
\contribution[\dagger]{Project Leader}

\abstract{In recent years, aerial object detection has been increasingly pivotal in various earth observation applications. However, current algorithms are limited to detecting pre-annotated categories. In this paper, we put forth a novel formulation of the aerial object detection problem, namely open-vocabulary aerial object detection (OVAD), which can detect objects beyond training categories without costly collecting new labeled data. We propose CastDet, a CLIP-activated student-teacher detection framework that serves as the first OVAD detector specifically designed for the challenging aerial scenario, where objects exhibit weak appearance features and arbitrary orientations. Our framework integrates a robust localization teacher along with several box selection strategies to generate high-quality proposals for novel objects. Additionally, the RemoteCLIP model is adopted as an omniscient teacher, which provides rich knowledge to enhance classification capabilities for novel categories. A dynamic label queue is devised to maintain high-quality pseudo-labels during training. By doing so, the proposed CastDet boosts not only novel object proposals but also classification. Furthermore, we extend our approach from horizontal to oriented OVAD with tailored algorithm designs for bounding box representation and pseudo-label generation. Extensive experiments for both tasks on multiple aerial datasets demonstrate the effectiveness of our approach.}

\date{\today}
\checkdata[Code]{\url{https://github.com/VisionXLab/CastDet}}
\checkdata[Journal]{International Journal of Computer Vision (IJCV'26)}

\begin{document}
\maketitle

% 如果需要目录，取消下面的注释
% \newpage
% \tableofcontents
% \newpage

\section{Introduction}
Object detection in aerial images aims at localizing and categorizing objects on the earth's surface, which is fundamental to a wide range of remote sensing applications, such as urban planning, environmental surveillance, and disaster response~\cite{zhao2003car,reilly2010detection,sadgrove2018real}. Despite significant advancements in aerial object detectors powered by deep learning~\cite{7926624,ding2019learning,yang2021r3det,qian2022rsdet++,yang2019scrdet,yang2019clustered}, they are still weak at detecting objects beyond the training categories. 

\begin{figure*}[t]
    \centering
    \begin{minipage}[t]{0.45\linewidth}
        \centering
        \includegraphics[width=\linewidth]{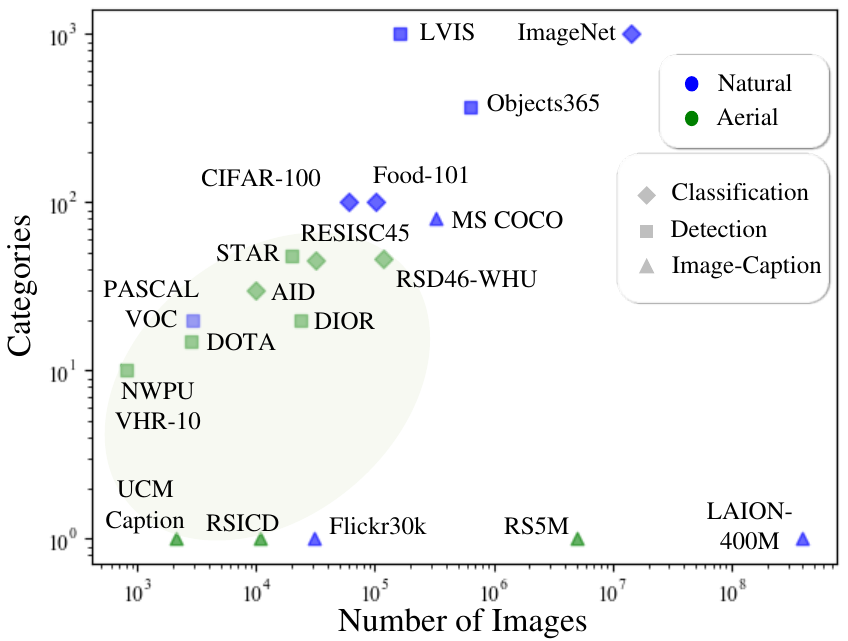}
        \caption{Comparisons of the amount of categories and images of 19 common aerial and natural image datasets. Aerial datasets are much smaller in size and category vocabularies compared with natural image datasets.}
        \label{fig:dataset_scale}
    \end{minipage}
    \hfill
    \begin{minipage}[t]{0.5\linewidth}
        \centering
        \includegraphics[width=0.9\linewidth]{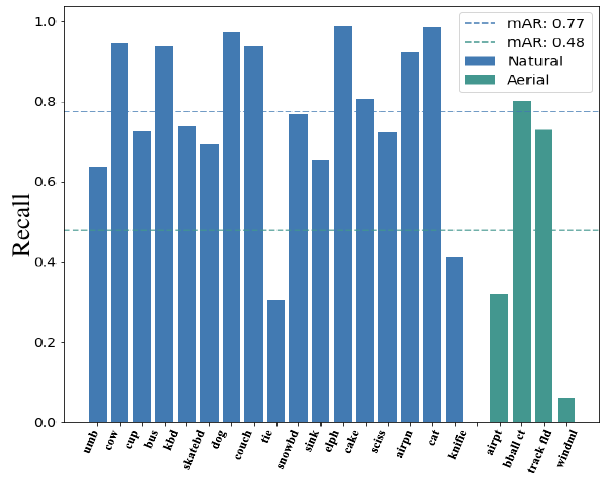}
        \caption{Class-agnostic RPN recall statistics of novel categories in the natural dataset COCO~\cite{lin2014microsoft} and the aerial dataset VisDroneZSD~\cite{VisDrone2023} (i.e., 77\% v.s. 48\%). The recall of novel objects in aerial images is much lower than that in natural images due to the highly complex backgrounds.}
        \label{fig:dataset_compare_recall}
    \end{minipage}
    \vspace{-10pt}
\end{figure*}

To expand the object detectors to novel categories, a naive idea is to collect and annotate large-scale aerial images with rich object categories. However, obtaining accurate yet sufficient annotations is time- and labor-expensive, even requiring human experts involvement. As a result, despite extensive efforts paid in the collection of existing aerial object detection datasets~\cite{nwpu_cheng2016learning, dior_li2020object, DOTA8578516, levir_zou2017random, visdrone_zhu2021detection,li2024star}, they are still much smaller and less diverse than natural image datasets~\cite{lin2014microsoft,gupta2019lvis,5206848imagenet}, as depicted in Fig.~\ref{fig:dataset_scale}. It hinders the scalability of detectors in open-world scenarios. 
Therefore, this paper advocates more flexible object detectors capable of detecting novel object categories unseen during training, referred to as open vocabulary object detection (OVD). This allows us to characterize new objects that emerged in the earth observation data without extra annotations in open scenarios.

\begin{figure*}[t]
  \centering
\includegraphics[width=\linewidth]{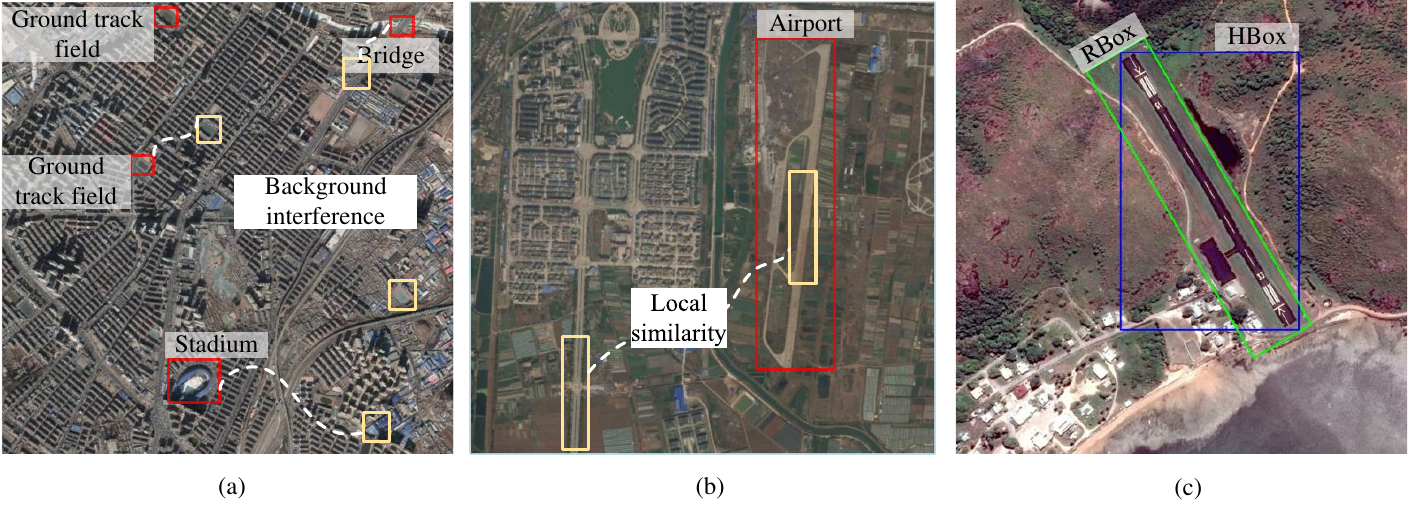}
   \caption{ Aerial images from DIOR~\cite{dior_li2020object}. (a)(b) Objects in aerial images exhibit background interference. The \textcolor{red}{red} box represents the ground truth, while the \textcolor{yellow}{yellow} box highlights the areas of interference. (c) Inaccurate localization with horizontal bounding box (HBox), causing significant background clutter in aerial images compared to oriented bounding box (RBox).}
   \label{fig:aerial_dataset_feature}
   \vspace{-10pt}
\end{figure*}

Drawing inspiration from the recent success of OVD in natural images~\cite{9879567glip, yao2022detclip, liu2023groundingdino}, we intend to explore ways in the challenging task, namely open vocabulary aerial object detection (OVAD), where objects often vary widely in scale, orientation, and exhibit weak feature appearance~\cite{Zhang2023}. 
Unlike natural images, which often feature clear contours and textures, allowing class-agnostic region proposal networks (RPNs) to effectively generalize to novel categories~\cite{chen2023ovarnet, zhou2022detic}; aerial images captured from overhead perspectives usually display weak surface features, where objects may blend into the background, as shown in Fig.~\ref{fig:aerial_dataset_feature}(a). This complicates object detection, for instance, AIRPORT may be confused with HIGHWAY, and thus be wrongly classified as background, as illustrated in Fig.~\ref{fig:aerial_dataset_feature}(b). Consequently, the recall of novel categories in aerial images is significantly lower than that in natural images, as demonstrated in Fig.~\ref{fig:dataset_compare_recall}. 
In addition, objects in aerial images exhibit great orientation diversity. Using horizontal bounding boxes for these rotated objects could bring in substantial background clutter, leading to inaccurate localization, as shown in Fig.~\ref{fig:aerial_dataset_feature}(c). However, current open-vocabulary detectors primarily support horizontal detection, which limits their effectiveness in aerial images.
These factors, on the one hand, pose challenges to directly apply current OVD methods for natural images to aerial images; and, on the other hand, spur us to develop extensible object detectors for aerial images, covering more object categories without extra annotation.

% para4: Solusion.
To address these challenges, we propose a simple yet effective open-vocabulary aerial detection framework, namely \textbf{CastDet}, a \textbf{C}LIP-\textbf{a}ctivated \textbf{s}tudent-\textbf{t}eacher \textbf{det}ector. It comprises  three key models: a student model and two teacher models for localization and classification, respectively. The student model is responsible for training both horizontal and oriented open-vocabulary detectors. 
The former adopts methods such as Faster R-CNN~\cite{ren2015faster}, and the later exploits the Oriented R-CNN~\cite{orientedrcnn_Xie_2021_ICCV}. 
The localization teacher model focuses on discovering and localizing potential objects, while the other teacher model provides classification guidance for novel categories. 
We jointly train the student model and teacher models in a self-learning manner, where the student model is optimized with pseudo-labels generated by the teachers, and the localization teacher model is updated from the continually learned student model. As exemplified by the literature~\cite{liu2021unbiased}, the slowly progressing localization teacher model acts as an ensemble of the previous student models, exhibiting superior capabilities that enable it to generate more reliable proposals. These proposals are then filtered based on their confidence scores, such as scale jittering variance, before being presented to the external teacher model.

While the student-teacher learning paradigm is powerful for enhancing the capabilities of the student model under the guidance of the teacher model~\cite{jeong2019consistency,sohn2020simple,zhou2021instant,liu2021unbiased,softteacher_xu2021end}, they operate in the closed-set setting. This leads to poor performance in discovering and recognizing novel object categories that are not encountered in the training data. To overcome this limitation, RemoteCLIP~\cite{liu2023remoteclip} that has been pre-trained on massive remote sensing image-text pairs and equips with strong zero-shot classification capabilities is exploited as an external teacher model. Thus, the learning process could benefit from the extra knowledge dedicated to interpreting aerial images brought by the RemoteCLIP. In addition to involving the RemoteCLIP, maintaining high-quality pseudo-labels is necessary. To this end, after predicting the category labels for the pseudo proposals, a dynamic label queue is proposed to store and iteratively update the pseudo-labels.

Unlike previous CLIP-based methods~\cite{gu2021open,zhong2022regionclip,chen2023ovarnet} that directly transfer knowledge from CLIP for open-vocabulary recognition, our CLIP-activated student-teacher self-learning framework incorporates high-confidence knowledge from RemoteCLIP as an incentive to guide both the student model and localization teacher model. This self-learning mechanism facilitates a ``flywheel effect'' wherein the external teacher model transfers knowledge to strengthen the localization teacher model to identify potential regions of novel objects, thereby enhancing pseudo box quality. In turn, the refined pseudo boxes enable the external teacher model to produce more precise pseudo-labels. Through our student-teacher self-learning process, the detection model can be progressively updated to localize and recognize continuously expanded object category vocabularies.

The preliminary content of this paper has partially appeared in the conference version, with the detector named CastDet~\cite{castdet} (ECCV 2024). The overall contributions of this extended journal version are as follows\footnote{This journal version extends the previous conference version (ECCV 2024)~\cite{castdet}, particularly in the following aspects: \textbf{1)} We rewrite the full paper with comprehensive discussions and more details; \textbf{2)} The proposed CastDet framework is now extended to support open-vocabulary oriented detection, addressing the specific challenges of detecting objects with arbitrary orientations in aerial scenes; \textbf{3)} We introduce two new oriented box selection strategies: box scale jittering variance and angle jittering variance, which take both the scale and orientation of objects into account to filter robust pseudo boxes; \textbf{4)} We integrate multiple open-vocabulary oriented detection algorithms into a unified code repository, including Oriented CastDet, Oriented ViLD, Oriented GLIP and Oriented GroundingDINO;  \textbf{5)} We conduct more ablation experiments to demonstrate the effectiveness of the proposed techniques;
\textbf{6)} We build a more comprehensive benchmark for the open-vocabulary aerial detection task, including both horizontal object detection and oriented object detection.}:

\begin{enumerate}
    \item Our work pioneers the open vocabulary aerial object detection (OVAD) to conquer fundamental challenges in interpreting earth observation images: the relatively small scale of annotated data in terms of both the number of training samples and the object categories; and the unique characteristics of aerial images, such as arbitrary orientations and weak feature appearances.
    \item We introduce the CastDet, a  flexible open-vocabulary detection framework incorporating the student-(multi)teacher self-learning paradigm. The proposed framework enables the detection model to progressively expand its object vocabulary and achieves accurate localization of novel objects using horizontal or oriented boxes without extra annotation efforts.
    \item We propose a series of box selection strategies to preserve high-quality pseudo-labels via filtering, including RPN Score, box jittering variance (BJV) and regression jittering variance (RJV) for horizontal boxes; scale jittering variance (SJV) and angle jittering variance (AJV) for oriented boxes.
    \item We devise a dynamic label queue for storing and incrementally updating high-quality pseudo-labels generated by the teacher models. By this way, richer and more accurate labels could be dynamically maintained.
    \item We implement multiple open-vocabulary oriented object detection methods, including Oriented ViLD, Oriented GLIP and Oriented GroundingDINO. This marks the first effort to open-vocabulary oriented detection.
    \item We utilize several public aerial image datasets to establish the open-vocabulary aerial detection benchmark and conduct extensive experiments to validate our methods, achieving significant performance improvement.
\end{enumerate}

\section{Related work}
\subsection{Close-set Object Detection}
Object detection aims to predict the bounding box coordinates and corresponding category labels for objects of interest within an image. Generally, existing object detection methods can be divided into two-stage and one-stage methods. Two-stage methods~\cite{rcnn_girshick2014rich, fastrcnn_girshick2015fast,ren2015faster,cascadercnn_cai2018cascade, maskrcnn_he2017mask} such as Faster R-CNN~\cite{ren2015faster} initially employ a region proposal network to generate potential target regions, then classify and refine these proposals. One-stage methods model the object localization and categorization as an end-to-end problem, which is to be resolved through a single network. Classical methods include anchor-based ones, such as YOLO~\cite{yolo_redmon2016you}, SSD~\cite{ ssd_liu2016ssd}, RetinaNet, \cite{retina_lin2017focal}, and anchor-free methods, e.g., CornerNet~\cite{law2018cornernet}, FCOS~\cite{tian2019fcos}, CenterNet~\cite{zhou2019objects}.

Inspired by the advanced progress of object detection methods~\cite{ren2015faster,retina_lin2017focal,yolo_redmon2016you,carion2020end,Zhang2023} dedicated to natural images, efforts have been paid to adapt these methods to aerial images. Specifically, several challenges remains to be tackled in aerial images, including object orientation~\cite{ding2019learning,orientedrcnn_Xie_2021_ICCV,han2021ReDet,yang2021r3det,qian2022rsdet++}, scale variation~\cite{10509806_glh_bridge,ZHENG20201_HyNet}, and dense objects~\cite{Yang2018AutomaticSD,yang2019scrdet,yang2019clustered,xu2022rfla}. Early works~\cite{7926624,zhang2016s,kang2017modified,liu2017sar} mainly focus on horizontal detection using classical deep learning methods to detect aerial objects with horizontal bounding boxes. 
However, horizontal bounding box could cause inaccurate localization of aerial objects with arbitrary orientations. 
Therefore, methods that could represent objects by oriented bounding boxes~\cite{Yang2018AutomaticSD,azimi2018towards,lee2023hausdorff} have attracted widespread attention. One focus of research is on optimizing the representation of oriented objects, which can be achieved by converting the angle regression task into an angle classification task~\cite{yang2022arbitrary,yang2021dense} or by modeling oriented objects using a Gaussian distribution~\cite{yang2021rethinking,yang2021learning,yang2022detecting}. In parallel, other studies emphasize learning robust alignment feature for oriented objects, exemplified by approaches such as SCRDet~\cite{yang2019scrdet}, R$^3$Det~\cite{yang2021r3det}, S$^2$A-Net~\cite{han2021align}, and ReDet~\cite{han2021ReDet}.

% ovd: natural; aerial (hbb, obb).
\subsection{Open-vocabulary Object Detection}
Open-vocabulary object detection (OVD) aims to detect objects beyond the categories known in the training phase. OVR-CNN~\cite{zareian2021open} pioneered this approach by combining bounding box annotations for a subset of categories with a corpus of image-caption pairs to extend the vocabulary of detectable objects.
Leveraging the outperforming zero-shot capabilities of pre-trained Vision-Language Models (VLMs), such as CLIP~\cite{CLIP_radford2021learning}, recent OVD methods success in achieving flexible and versatile detection with extensible object categories by means of prompt learning~\cite{du2022learning, feng2022promptdet,zang2022open} or region-level fine-tuning~\cite{gu2021open,zhong2022regionclip,zhou2022detic,wu2023cora}. For instance, ViLD~\cite{gu2021open} transfers knowledge from the CLIP to enrich a two-stage detector via both vision and language knowledge distillation. RegionCLIP\cite{zhong2022regionclip} aligns visual representations at the region level with text descriptions. Detic~\cite{zhou2022detic} enriches the object vocabulary by incorporating image classification data to expand the scope of detectable objects to tens of thousands. PromptDet~\cite{feng2022promptdet} and DetPro~\cite{du2022learning} take prompt learning techniques to enhance the alignment between visual feature and semantic feature at the region level. Additionally, some works~\cite{9879567glip,zhang2022glipv2,liu2023groundingdino} pre-train the detector capable of open-vocabulary detection with unified object detection and phrase grounding objectives. For example, GLIP~\cite{9879567glip} integrates image-language pre-training to align visual and textual data, improving object detection across diverse categories using text prompts. GroundingDINO~\cite{liu2023groundingdino} combines the DINO detector with grounded pre-training on large-scale datasets to enhance open-set object detection by the tight fusion between vision and language.
The success of these approaches relies on two critical factors: 1) effective object proposal generation beyond training categories; and 2) large-scale image-text datasets that enable robust open-vocabulary classification capabilities.

The limited size of existing aerial image datasets, along with their distinct visual characteristics compared to natural images, leads to unsatisfying generalization in region proposal generation for diverse categories. Consequently, open-vocabulary detection methods developed for natural images perform poorly on aerial images. To address the issue in open-vocabulary aerial detection, we introduce the CastDet~\cite{castdet}, the first approach dedicated to tackle the unique challenges posed by aerial imagery through a student-multi-teacher self-learning paradigm. In addition, a few studies have been exploring the OVAD following the CastDet. For example, LAE-DINO~\cite{pan2024locate} proposes a DINO-based detector with dynamic vocabulary construction and visual-guided text prompt learning to improve detection across diverse aerial categories. OVA-DETR~\cite{wei2024ovadetropenvocabularyaerial} enhances feature extraction by leveraging image-text alignment with region-text contrastive loss and Bidirectional Vision-Language Fusion, particularly for small objects. However, none of these methods address the challenge of oriented object detection in aerial scenes. To this end, another goal of this paper is to fill this blank and set a foundation for future research.

\section{Methodology}
We start by presenting the open-vocabulary object detection problem setting in Section~\ref{subsec:preliminaries}). Next, we provide the overview of our CastDet framework in Section \ref{subsec:ovd}, which is composed of three key components: 
\rmnum{1}) the student module, a two-stage open-vocabulary detector, introduced in Section \ref{sec:student}; 
\rmnum{2}) the localization teacher model, as detailed in Section \ref{sec:localization_teacher}; 
and \rmnum{3}) the dynamic pseudo-label queue for maintaining high-confidence pseudo-labels, outlined in Section \ref{sec:dynamic_queue}. 
Finally, we elaborate on two tasks: horizontal open-vocabulary detection in Section~\ref{sec:hbb_ovd} and oriented open-vocabulary detection in Section~\ref{sec:obb_ovd}.

\subsection{Definition of Open Vocabulary Aerial Object Detection}
\label{subsec:preliminaries}

Open Vocabulary Aerial Object Detection (OVAD) aims to extend traditional aerial object detection within base classes to detect novel objects emerged in earth observation, addressing the unique challenges inherent to aerial scenarios, including weak feature appearance, orientation diversity, and the difficulty of obtaining sufficient labeled data. Given an image $\mathbf{x}\in \mathbb{R}^{H\times W\times 3}$ , the goal of the newly defined OVAD task is to train a detector capable of detecting both base ($\mathcal{C}_{\text{base}}$) and novel ($\mathcal{C}_{\text{novel}}$) objects within the image. This task involves solving two subproblems: \rmnum{1}) localize all objects with precise bounding boxes, where each box is represented by coordinates  $b_{i}\in \mathbb{R}^4$  for horizontal objects, and  $b_{i}\in \mathbb{R}^5$ for oriented objects; and \rmnum{2}) classify the $i$-th object with the category label $c_{i}\in \mathbb{R}^{\mathcal{C}_{\text{test}}}$, where  $\mathcal{C}_{\text{test}}=\mathcal{C}_{\text{base}}\cup\mathcal{C}_{\text{novel}}$ and  $\mathcal{C}_{\text{base}}\cap\mathcal{C}_{\text{novel}}=\emptyset$. During training, we utilize a labeled dataset $\mathcal{L}=\{(\mathbf{x},\{(b,c)_k\})_i\}_{i=1}^{|\mathcal{L}|}$ on a set of base categories, i.e., $c_{k}\in \mathbb{R}^{\mathcal{C}_{\text{base}}}$. Additionally, we incorporate an unlabeled dataset  $\mathcal{U}=\{\mathbf{x}_i\}_{i=1}^{|\mathcal{U}|}$, which consists of images without bounding box annotations. The combined training dataset is represented as $\mathcal{D}_{\text{train}}=\mathcal{L}\ \cup\ \mathcal{U}$. 

\subsection{CLIP-Activated Student Teacher Framework}
\label{subsec:ovd}
\begin{figure*}[t]
  \centering
   \includegraphics[width=\linewidth]{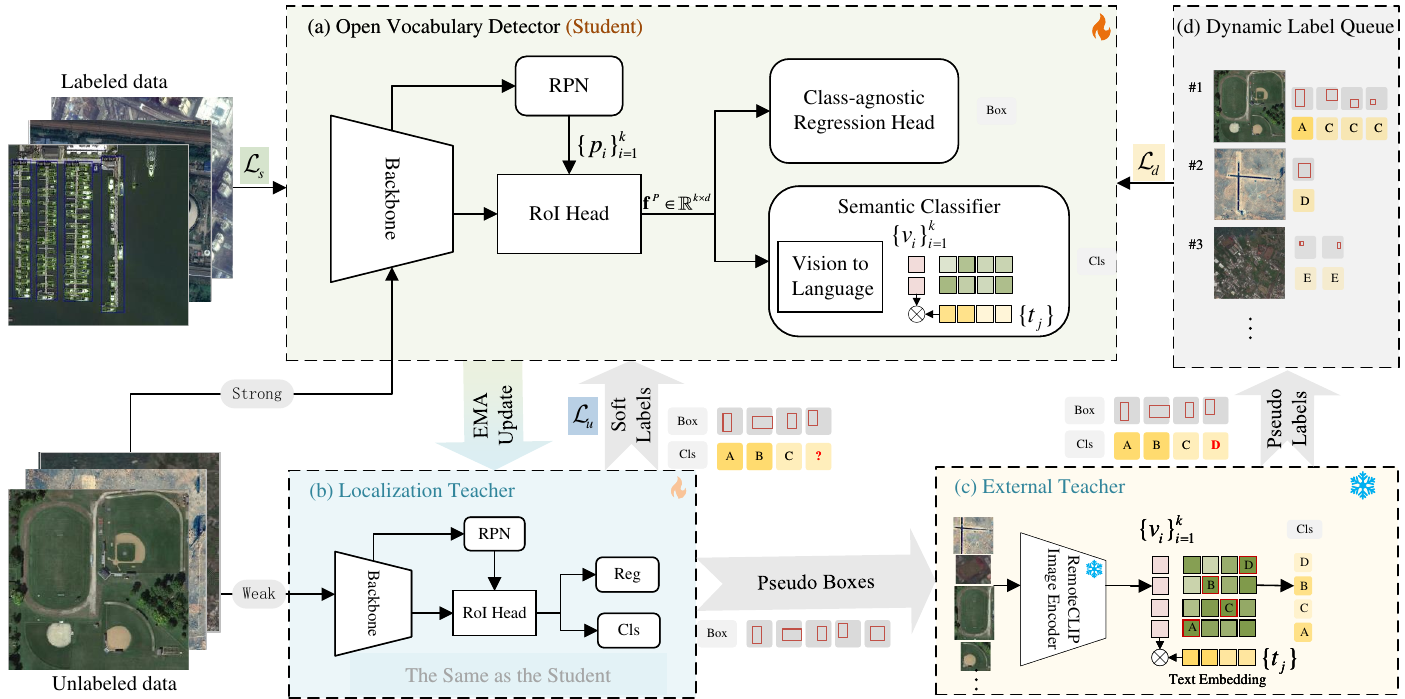}
   \caption{Overall architecture of CastDet. In each training iteration, the data batch consists of three data flow: labeled data with annotations, unlabeled data, and data sampled from the dynamic label queue. The labeled images are directly used for the student model training (\colorbox[rgb]{0.84,0.894,0.805}{$\mathcal{L}_s$}), while two sets of pseudo-labels of unlabeled data are predicted through the localization teacher model and external teacher model. One supervises the student model (\colorbox[rgb]{0.726,0.816,0.898}{$\mathcal{L}_u$}), and the other is pushed into the dynamic label queue. Simultaneously, samples are randomly selected from the dynamic label queue to enhance the student's ability to detect novel objects (\colorbox[rgb]{0.99,0.94,0.805}{$\mathcal{L}_d$}).}
   \label{fig:framework}
\end{figure*}

\textbf{Architecture Overview.} Fig. \ref{fig:framework} illustrates the overview of our CastDet framework, which consists of a student model and two teacher models: a localization teacher and an external teacher. The student model is a two-stage object detection model (e.g., Faster R-CNN or Oriented RCNN) with a modified class-agnostic bounding box regression head ($\Phi_{\text{LOC}}$) and a semantic classifier ($\Phi_{\text{CLS}}$).  It is trained on both labeled and unlabeled samples with pseudo boxes and category labels generated by the localization teacher model and external teacher model. The localization teacher model is an exponential moving average (EMA) of the student model~\cite{tarvainen2017mean}, which can aggregate the history information during the training iterations to obtain better and stable representations, ensuring the quality of the pseudo-labels. During training, the localization teacher model generates two sets of pseudo-labels for the unlabeled images, one is used for training the student model, and another serves as input to the external teacher model for generating pseudo category labels. The external teacher model is a frozen RemoteCLIP~\cite{liu2023remoteclip}, a vision-language fundamental model pre-trained on large-scale remote sensing image-text pairs, following the CLIP framework. It provides strong open-vocabulary classification ability by comparing the similarity between image and category embeddings. Additionally, a dynamic label queue is used to store the pseudo-labels generated by the external teacher model, facilitating to maintain high-quality pseudo-labels and transfer data samples for the student model training.

\subsubsection{Student}
\label{sec:student}
The student model is a two-stage object detector, represented by Faster R-CNN or Oriented-RCNN, comprising a backbone encoder ($\Phi_{\text{ENC}}$), a region proposal network ($\Phi_{\text{RPN}}$), and a box head module ($\Phi_{\text{BOX}}$):
\[
\{\hat{y}_1, \ldots, \hat{y}_n\} = \Phi_{\text{BOX}} \circ \Phi_{\text{RPN}} \circ \Phi_{\text{ENC}}(\mathbf{x}),
\]
where the operator ``$\circ$'' denotes function composition, i.e., the input $\mathbf{x}$ is processed sequentially by $\Phi_{\text{ENC}}$, $\Phi_{\text{RPN}}$, and $\Phi_{\text{BOX}}$.

To construct an open-vocabulary detector capable of detecting novel objects beyond the close-set categories, we incorporate two key components into the box head:
\rmnum{1}) a class-agnostic box regression head ($\Phi_{\text{LOC}}$) to generate class-agnostic box coordinates, and \rmnum{2}) a semantic classifier head ($\Phi_{\text{CLS}}$) to classify these objects beyond seen categories, i.e., open-vocabulary classification.

\textbf{Class-agnostic box regression head ($\Phi_{\text{LOC}}$)} takes the RoI (Region of Interest) features as input and outputs refined box coordinates for each object. To handle a variable number of object categories, $\Phi_{\text{LOC}}$ shares parameters across all categories. Specifically, the regression box is represented as $b_i\in \mathbb{R}^k$ instead of $b_i\in\mathbb{R}^{k|\mathcal{C}_{\text{test}}|}$ for each box $i$, where $k\in \{4,5\}$ is the dimension of the box, with $k=4$ for horizontal boxes and $k=5$ for oriented boxes. This approach is also in line with the previous works~\cite{feng2022promptdet,zhou2022detic}.

\textbf{Semantic classifier head ($\Phi_{\text{CLS}}$)} is devised for classifying RoI regions beyond a predefined set of categories. Following Detic\cite{zhou2022detic}, we employ semantic embeddings for category vocabularies as the weights of the last fully connected layer for flexible category expansion. The semantic embeddings are generated by two steps: \rmnum{1}) filling concept into the predefined prompt template  ``\texttt{a photo of [\underline{category}]}'', and \rmnum{2}) extracting semantic embeddings of text descriptions using the text encoder of RemoteCLIP.
\begin{equation}
    t_j=\Phi_{\text{clip}}^{\text{text}}  \circ \Phi_{\text{clip}}^{\text{tokenizer}}( \text{a photo of [category]}),
\end{equation}
where $\Phi_{\text{clip}}^{\text{tokenizer}}(\cdot)$ denotes the text tokenizer that maps text into tokens, $\Phi_{\text{clip}}^{\text{text}}(\cdot)$ is the text encoder that converts tokens into the final semantic embedding $t_j$ for category $j$.

Given a set of RoI visual features $\{v_i\}$, the similarity score between $i$-th RoI feature and $j$-th category is calculated as:
\begin{equation}
\label{equ:s_ij}
    \hat{s}_{ij}=\frac{v_i^T\cdot t_j}{\tau\left \| v_i \right \|\cdot \left \| t_j \right \|},
\end{equation}
where $\tau$ is the temperature parameter that controls the range of the logits, and it is optimized as a log parameterized multiplicative scalar as in \cite{CLIP_radford2021learning}.

\begin{figure*}[t]
  \centering
   \includegraphics[width=\linewidth]{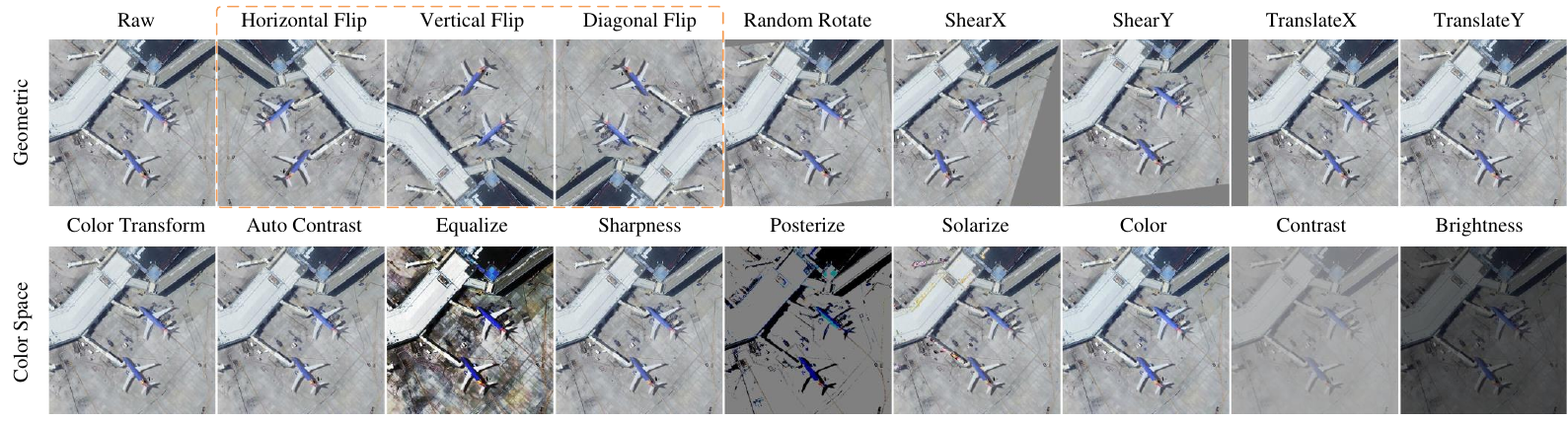}
   \caption{Augmentation strategies used in our approach.Weak augmentation applies only random flip (highlighted in \textcolor{orange}{orange} box), whereas strong augmentation incorporates all augmentation techniques.}
   \label{fig:vis_aug}
\end{figure*}

\subsubsection{Localization Teacher}
\label{sec:localization_teacher}
The quality of region proposals is significant for effectively detecting novel objects, as it directly impacts the subsequent box refinement and classification. However, as shown in Fig.~\ref{fig:dataset_compare_recall}, the RPN recall of novel categories in aerial images is significantly lower than that in natural images. To address this problem, we employ a robust teacher model to generate better pseudo-boxes for training the student model. Successively, deriving a better teacher model from the student one without introducing additional training costs is crucial. A naive solution is to employ a frozen teacher model trained on base categories, however, a issue rises that it lacks flexibility in learning novel categories in the open-vocabulary object detection. Thus, we opt for a continuously updating mechanism that could enable the teacher model to evolve and adapt to novel concepts, thereby enhancing its effectiveness in a broader detection scenarios.

\textbf{Exponential Moving Average.} Inspired by the literature~\cite{tarvainen2017mean,  he2020momentum}, we exploit a momentum update strategy for optimizing teacher model, that is, the teacher model is updated by an exponential moving average of the student model during training.  Formally, denoting the parameters of the teacher model as $\theta^\prime$ and those of the student model as $\theta$. Only the parameters $\theta$ are updated via back-propagation, and parameters $\theta^\prime$are updated as a weighted average of successive weights of $\theta$ at each training iteration $t$:
\begin{equation}
\label{equ:ema}
    \theta_t^\prime = \alpha\theta_{t-1}^\prime + (1-\alpha)\theta_t, 
\end{equation}
where $\alpha\in [0,1)$ is a momentum coefficient.

The momentum update described in Equ.~(\ref{equ:ema}) enhances the teacher model in several significant ways compared to a frozen teacher model:  1) it accumulates historical information of the student model without additional training costs, thereby obtaining better intermediate representations and more robust predictions~\cite{tarvainen2017mean,liu2021unbiased}; 2) it enables the evolving teacher model to leverage unlabeled data effectively, enhancing the performance of the model with fewer annotated images; and 3) it facilitates on-line learning, allowing the model to continuously adapt and scale to an increasing number of novel concepts.

\textbf{Consistency Training.}
Following \cite{softteacher_xu2021end,sohn2020fixmatch}, we utilize different augmentation strategies for pseudo labeling. As shown in Fig.~\ref{fig:vis_aug}, weak augmentation involves only random flip, as highlighted by the orange box in the image. On the other hand, strong augmentation incorporates a wide range of augmentation techniques, including both geometric and color space augmentations such as Random Rotate, Auto Contrast, and others.

We adopt Consistency training~\cite{jeong2019consistency,sohn2020simple} to guide the student model to make the same prediction as the teacher model makes. Specifically, given an unlabeled image $\mathbf{x}$, it first goes through weak augmentation and serves as the input for the localization teacher model to generate pseudo-labels. The pseudo-labels are then used to supervise the student model,  which takes the same image with a strong augmentation as input. Formally, the teacher model's prediction is defined as $\mathbf{\hat{y}}=f(T^\prime(\mathbf{x}),\theta^\prime)$. The objective is to minimize the consistency cost $J$  between $\mathbf{\hat{y}}$  and the predictions of the student model, i.e.,
\begin{equation}
    J(\theta) = \mathcal{L}(f(T(\mathbf{x}),\theta),\mathbf{\hat{y}}), 
\end{equation}
where $T$ and $T^\prime$ denote strong and weak augmentation, respectively. The cost function $J$ is optimized using stochastic gradient descent (SGD). A detailed explanation of the loss function is provided in Sec.~\ref{sec:hbb_hybrid_training}.

\subsubsection{Dynamic pseudo-label Queue}
\label{sec:dynamic_queue}
\begin{figure*}[t]
  \centering
   \includegraphics[width=0.95\linewidth]{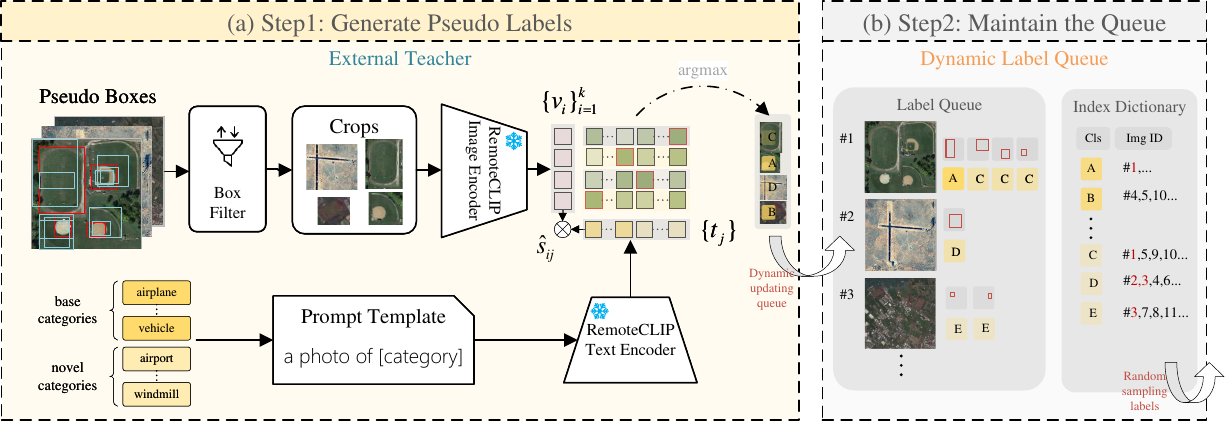}
   \caption{Workflow of the dynamic label queue. \textbf{Step1:} filter certain high-quality proposal boxes generated by the localization teacher model, and employ the RemoteCLIP to classify corresponding cropped images as pseudo-labels. \textbf{Step2:} dynamically update the pseudo-labels into the queue, and randomly sample a batch of pseudo-labels for the student model training.}
   \label{fig:dynamic_queue}
\end{figure*}
The dynamic label queue is devised for storing, dynamically updating, and transferring pseudo-labels. At each training iteration, the localization teacher model and external teacher model generate high-quality pseudo-labels for the unlabeled images, which are then pushed into the label queue. Following this, a random batch of pseudo-labels is sampled from the dynamic queue for the student model training, as illustrated in Fig.~\ref{fig:dynamic_queue}.

\textbf{Generate pseudo-labels.}
The localization teacher model takes unlabeled images with weak augmentation $T^\prime(\mathbf{x})$ as input, where the backbone first extracts features, the RPN then generates a set of proposals $\{p_i\}$ and extracts RoI features $\mathbf{f}^P= \Phi_{\text{RPN}} \circ \Phi_{\text{ENC}}(T^\prime(\mathbf{x}))$ of these proposals through RoI pooling. Subsequently, the class-agnostic regression branch predicts the refined coordinates  $\{\hat{b}_i\}$. To improve the accuracy of pseudo-labels while reducing computational complexity, we use a box filter to select $k$ candidates and obtain a set of image crops $\{\mathbf{x}^I_i\}_{i=1}^{k}$ by cropping the corresponding regions of the image. The category labels for these regions are then predicted by the RemoteCLIP.

For each cropped image $\mathbf{x}^I_i$, the visual feature $v_i$ is computed via the visual encoder of the RemoteCLIP, denoted as $ v_i=\Phi_{\text{clip}}^{\text{visual}}(\mathbf{x}^I_i)$. At the same time, text embeddings $t_j$ corresponding to category descriptions are generated, as detailed in Sec.~\ref{subsec:ovd}. The prediction probability for each category $j$ given the cropped image $i$ is the softmax value for similarity between the visual feature and text embeddings:
\begin{equation}
    \hat{p}_{ij}=\frac{e^{\hat{s}_{ij}}}{\sum_k e^{\hat{s}_{ik}}},
\end{equation}
where $\hat{s}_{ij}$ is the similarity score, calculated by Equ (\ref{equ:s_ij}).

We further apply a threshold $p_0$ to filter pseudo-labels,

\begin{equation}
    \begin{aligned}
        \hat{y} &= \left\{ (\hat{b}_i, \hat{c}_i) \mid \hat{p}_i \ge p_0 \right\}_{i=1}^{\hat{k}}, \\
        \hat{c}_i &= \arg\max_j \hat{p}_{ij},\ \hat{p}_i = \max_j \hat{p}_{ij},
    \end{aligned}
\end{equation}
where $\hat{p}_i$, $\hat{c}_i$ are the prediction score and category label, respectively. Finally, the image with its pseudo-labels $(\mathbf{x},\hat{y})$ is pushed into the dynamic label queue.

\textbf{Maintain the Queue.}
As shown in Fig.~\ref{fig:dynamic_queue}(b), the dynamic label queue holds both image metadata and the associated pseudo-labels generated by the teacher models. Image metadata includes details such as the image index and file path, which are critical for tracking and retrieving images during the training process. Pseudo-labels, on the other hand, are comprised of annotations such as bounding boxes and class labels. To efficiently manage this queue, we utilize an index dictionary to store the mapping relationship between categories and image indexes, represented  as \texttt{\{cls\_id:list[image\_ids]\}}. As new images are processed and pseudo-labels are generated, they are added to the queue, and the index dictionary is adjusted to include these updates. 
Additionally, images with pseudo-labels are sampled with a predefined probability based on the index dictionary, which are then used to train the student model.

This dynamic process allows the queue to progressively accumulate richer and more accurate pseudo-labels as the model iterates, with visualization results shown in Fig.~\ref{fig:queue_iter_analysis}.
The continuous improvement benefits the entire system, enabling the student and localization teacher models to learn from diverse, high-quality pseudo-labels. 

\subsection{Horizontal Open-vocabulary Detection}
\label{sec:hbb_ovd}
Horizontal object detection aims to detect all objects in a image and represent their location with horizontal bounding boxes, i.e., $b_i\in \mathbb{R}^4$. 
To enable open-vocabulary detection, we build upon the Faster R-CNN framework~\cite{ren2015faster} and incorporate external knowledge from unlabeled images, which may contain information about novel categories. Specifically, the localization teacher model first generates a set of box candidates, which are then filtered based on their confidence scores. Only those with high confidence are forwarded to the external teacher model for category prediction. The model is then optimized using a hybrid training mechanism that leverages both labeled data and unlabeled data with these high-quality pseudo-labels.

\subsubsection{Horizontal Box Selection Strategies}
\begin{figure*}[!t]
  \centering
   \includegraphics[width=0.98\linewidth]{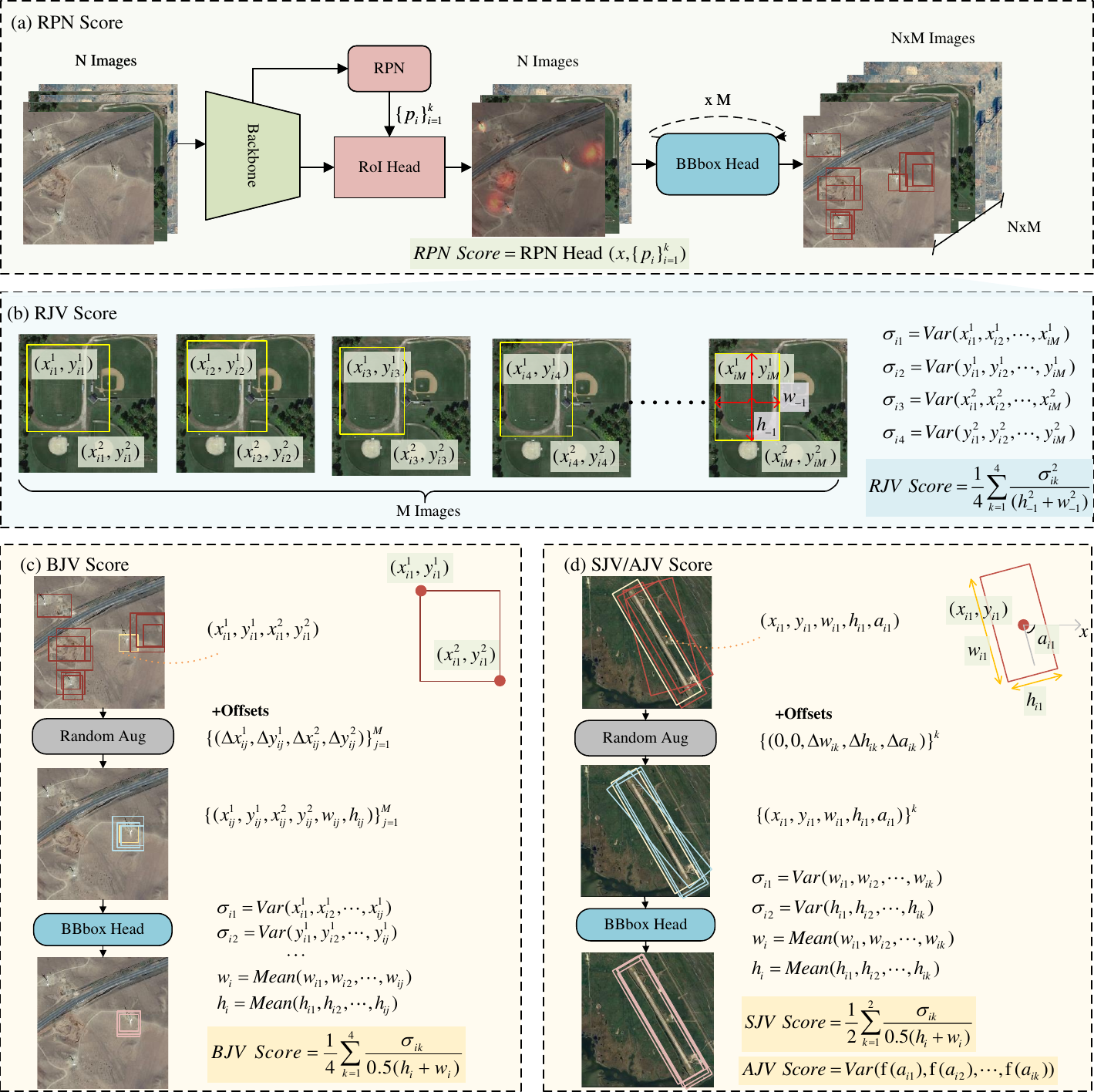}
    \caption{Illustration for 5 types of box selection strategies, including horizontal box selection strategies: (a) RPN Score, (b) regression jittering score (RJV), (c) box jittering score (BJV); and oriented box selection strategies: (d) scale jittering score (SJV) and angle jittering score (AJV).}
    \label{fig:hbox_rbox_selector}
\end{figure*}

The quality of pseudo-boxes is vital, as it directly affects the accuracy of the category labels generated by the external teacher model, subsequently influencing the overall performance of the model. Given this context, we utilize several box confidence scores to filter high-quality pseudo boxes, including RPN, RJV and BJV Score, as shown in Fig.~\ref{fig:hbox_rbox_selector}(a)$\sim$(c).

\textbf{RPN Score.}
The RPN Score represents the confidence that a region contains a object. This strategy filters out boxes with low foreground confidence score, which is a common approach adopted by most OVD methods~\cite{zhao2022exploiting,chen2023ovarnet}.  As Shown in Fig.~\ref{fig:hbox_rbox_selector}(a), the RPN Score can be directly obtained from the region proposal network:
\begin{equation}
    \text{RPN Score}=\Phi_{\text{RPN}}^{\text{score}} \circ \Phi_{\text{ENC}}(T^\prime(\mathbf{x})).
\end{equation}

\textbf{Regression Jittering Variance (RJV).}
The RJV Score reflects the stability of object location predictions across $M$ regression steps. To achieve a more precise localization, we employ regression jittering, which iteratively refines the predicted box by feeding it back into the regression branch ($\Phi_{\text{LOC}}$) for multiple steps. Specifically, given an initial predicted box, we iteratively pass this box through the regression branch $M$ times. Each iteration outputs a refined bounding box prediction, where slight variations in the coordinates may occur. The RJV Score quantifies this instability by measuring the variance of the predicted coordinates across these regression steps.
Formally, the RJV Score is defined as 
\begin{equation}
    \bar{\sigma_i} = \frac{1}{4}\sum_{k=1}^4 \frac{\sigma_{ik}^2}{(h_{-1}^2+w_{-1}^2)},
\end{equation}
where $\{\sigma_{ik}\}_{k=1}^4$ represents the standard derivation of the four box coordinates (i.e., $x_1$, $y_1$, $x_2$, $y_2$) for the $i$-th set of regression boxes obtained across $M$ iterations. The denominators $h_{-1}$ and $w_{-1}$ denote the height and width of the final predicted box after $M$ steps, which serve to normalize the variance relative to the size of the box, thus making the score scale-invariant.

\textbf{Box Jittering Variance (BJV).}
The BJV Score is employed to measure the stability of object location predictions across multiple jittered boxes. Box jittering involves introducing random noise around a predicted bounding box $b_i$ to generate $M$ jittered boxes. These jittered box are then processed by the box regression branch ($\Phi_{\text{LOC}}$) to predict the refined bounding boxes $\{\hat{b_{i,j}} \}$, where $j$ ranges from 1 to $M$. The BJV is defined as follows
\begin{equation}
 \bar{\sigma_i} = \frac{1}{4}\sum_{k=1}^4 \frac{\sigma_{ik}}{0.5(h_i+w_i)},
\end{equation}
where $\{\sigma_{ik}\}_{k=1}^4$ represents the standard derivation of the coordinates (i.e., $x_1$, $y_1$, $x_2$, $y_2$) for the $i$-th set of refined boxes. The terms $h_i$, $w_i$ denote the mean height and width of the $i$-th boxes set, respectively~\cite{softteacher_xu2021end}.

\subsubsection{Hybrid Training}
\label{sec:hbb_hybrid_training}
We follow the student-teacher training mechanism. In each training iteration, three types of data flow are sampled from the training batch: The labeled data is directly used to supervise the student model ($\mathcal{L}_s$). The unlabeled data is assigned pseudo-labels by the localization teacher model and contributes to the unsupervised loss ($\mathcal{L}_u$).  The pseudo-labeled data from the dynamic label queue encourages the model to discover novel objects ($\mathcal{L}_d$),  as depicted in Fig.~\ref{fig:framework}. Thus,  the overall loss comprises three components:
\begin{equation}
\label{equ:losses}
    \mathcal{L} = \alpha\mathcal{L}_s+ \beta\mathcal{L}_u + \gamma\mathcal{L}_d,
\end{equation}
where $\mathcal{L}_s$, $\mathcal{L}_u$ and $\mathcal{L}_d$ represent the supervised loss for labeled images, unsupervised loss for unlabeled images with pseudos generated by the localization teacher model, and unsupervised loss for images sampled from the dynamic label queue, respectively. The weights $\alpha$, $\beta$ and $\gamma$ balance these loss components.

\textbf{Labeled Data Flow.} 
To achieve horizontal object detection, our approach adopts the architecture of Faster R-CNN, adapted to an open-vocabulary framework. In a typical processing batch, we receive a batch of labeled data $\{(\mathbf{x}_k,\{(b_i,c_i)\})\}$. The student model predicts the bounding box coordinates $\{\hat{b}_i\}$ and associated prediction scores $\{\hat{s}_i\}$ for these images, as depicted in Fig.~\ref{fig:framework}(a). The supervised loss is then calculated as
\begin{equation}
        \mathcal{L}_s = \frac{1}{N_b}\sum_{i=1}^{N_b}  \mathcal{L}_{\mathrm{cls}}(\hat{s}_i,c_i)+\frac{1}{N_{b}^{\mathrm{fg}}}\sum_{i=1}^{N_{b}^{\mathrm{fg}}} \mathcal{L}_{\mathrm{reg}}(\hat{b}_i, b_i),
\end{equation}
where $\mathcal{L}_{\mathrm{cls}}$ is the classification loss, calculated using Cross Entropy Loss. $\mathcal{L}_{\mathrm{reg}}$ is the box regression loss, calculated using L1 Loss. $N_b$ denotes the total number of proposals generated by the model for each image, and $N_{b}^{\mathrm{fg}}$ represents the number of foreground proposals.

\textbf{Unlabeled Data Flow.}
The unsupervised loss $\mathcal{L}_u$ comprises two main components: the classification loss $\mathcal{L}_u^{\mathrm{cls}}$ and the box regression loss $\mathcal{L}_u^{\mathrm{reg}}$. During the initial training phases, the detector often struggles to correctly identify novel categories, leading to high rates of false negatives. To mitigate this issue without overly penalizing the model for incorrect background predictions, we employ a weighted mechanism for negative samples. The classification loss is defined as:
\begin{equation}
    \mathcal{L}_{u}^{\mathrm{cls}}=\frac{1}{N_{b}^{\mathrm{fg}}} \sum_{i=1}^{N_{b}^{\mathrm{fg}}} \mathcal{L}_{\mathrm{cls}}\left(\hat{s}_{i}, \hat{c}_i\right)+\sum_{j=1}^{N_{b}^{\mathrm{bg}}} w_{j} \mathcal{L}_{\mathrm{cls}}\left(\hat{s}_{j}, \hat{c}_j\right),
\end{equation}
where $N_{b}^{\mathrm{bg}}$ denotes the total number of background objects. The weight $w_j$ for each background sample is determined by the normalized contribution of the background prediction score of the $j$-th candidate. 

For training the box regression branch, we first apply the BJV score to filter candidates, as described in Sec.~\ref{sec:localization_teacher}. The regression loss is then defined as:
\begin{equation}
    \mathcal{L}_{u}^{\mathrm{reg}}=\frac{1}{N_{b}^{\mathrm{fg}}} \sum_{i=1}^{N_{b}^{\mathrm{fg}}} \mathcal{L}_{\mathrm{reg}}\left(\hat{b}_{i}^{\mathrm{fg}}, \hat{b}_i\right),
\end{equation}
where $\hat{b}_{i}^{\mathrm{fg}}$ and $\hat{b}_i$ denote the $i$-th predicted foreground box and the assigned pseudo box, respectively.

\textbf{Queue Data Flow.} 
To enhance the open-vocabulary capability, we randomly sample a batch of images from the dynamic label queue to train the student model, see Sec.~\ref{sec:dynamic_queue}. Like unsupervised loss, $\mathcal{L}_d$ also includes two parts: $ \mathcal{L}_{d}^{\mathrm{cls}}$ and $ \mathcal{L}_{d}^{\mathrm{reg}}$. Since the localization teacher model primarily guides the student model in discovering and localizing objects, the $ \mathcal{L}_{d}^{\mathrm{reg}}$ is optional in this process. The classification loss is defined as follows:
\begin{equation}
        \mathcal{L}_{d}^{\mathrm{cls}}=\frac{1}{N_{b}^{\mathrm{fg}}} \sum_{i=1}^{N_{b}^{\mathrm{fg}}} \mathcal{L}_{\mathrm{cls}}\left(\hat{s}_{i}, \hat{c}_i\right)+\sum_{j=1}^{N_{b}^{\mathrm{bg}}} w_{j} \mathcal{L}_{\mathrm{cls}}\left(\hat{s}_{j}, \hat{c}_j\right),
\end{equation}
where $w_j$ is a weighting factor, which balances the foreground and background loss.

\subsection{Oriented Open-vocabulary Detection}
\label{sec:obb_ovd}
Aerial objects often exhibit arbitrary orientation, the generic horizontal object detector cannot locate these oriented aerial objects accurately. Therefore, we develop oriented open-vocabulary detector to address this problem. We utilize the Oriented R-CNN~\cite{orientedrcnn_Xie_2021_ICCV} framework as the base detector, which is capable of representing the objects with their oriented bounding boxes, each box is denoted by a 5-dimensional vector  $b_i=(c_x,c_y,w,h,a)$, where $(c_x,c_y)$ is the center coordinate of the box, $w$ and $h$ represent the width and height of the box, and $a$ specifies the angle of rotation. The training objective are almost the same as horizontal detection training, as detailed in Sec.~\ref{sec:hbb_hybrid_training}. However, when generating oriented pseudo-labels, it is crucial to consider both the scale and orientation of the bounding boxes. This section describes two advanced strategies for selecting robust pseudo oriented boxes.

\subsubsection{Oriented Box Selection Strategies}
To filter robust pseudo boxes, we propose two types of rotated box selection strategies, box scale jittering variance (SJV) and angle jittering variance (AJV), which take scale and orientation of the objects into consideration, respectively.

\textbf{Box Scale Jittering Variance (SJV).}
The SJV Score evaluates the scale stability across multiple noised instances. We first randomly add noise $(\Delta w,\Delta h,\Delta a)$  to box representations $(w,h,a)$, thus generate $M$ jittered boxes. Here, we fix the center point $(c_x,c_y)$ for each box. Then, the noised boxes are refined by the oriented regression branch. Finally, the scale jittering variance is calculated by
\begin{equation}
     \bar{\sigma_i} = \frac{1}{2}\sum_{k=1}^2 \frac{\sigma_{ik}}{0.5(h_i+w_i)},
\end{equation}
where $\sigma_{ik}$ represents the standard deviation of the scale components (height and width), $h_i$ and $w_i$ are the mean height and width of the $i$-th set of jittered boxes.

\begin{figure}[!tb]
  \centering   \includegraphics[width=0.7\linewidth]{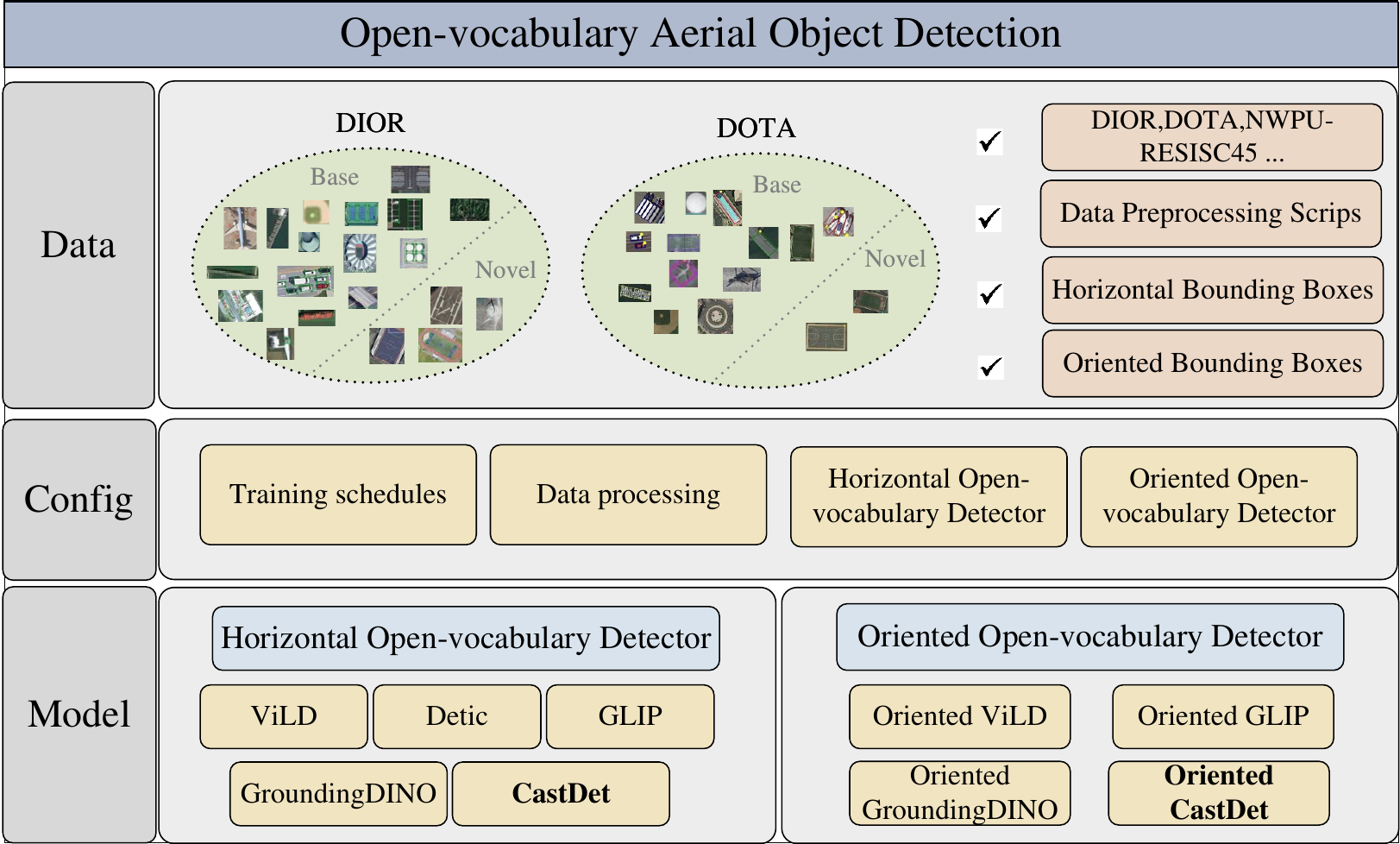}
   \caption{Overview for open-vocabulary aerial detection framework.}
   \label{fig:benchmark}
\end{figure}

\textbf{Angle Jittering Variance (AJV).}
The AJV is used to access the stability of angle predictions, we first generate $M$ jittered boxes follows the jittering strategy of SJV, and then compute angle jittering score by
\begin{equation}
\label{equ:ajv}
     \bar{\sigma_i} = \mathrm{Var}(\mathrm{f}(a_{i1}),\mathrm{f}(a_{i2}),\cdots,\mathrm{f}(a_{ik})),
\end{equation}
where $a_{ij}$ denotes the angle of $j$-th box of $i$-th box set. The function $\mathbf{f}(\cdot)$ denotes a function of the angle, which transforms the angle values to better capture the variance in orientation, e.g.,  $\sin (\cdot)$ or $\text{Identity}(\cdot)$ function. Here, $\mathrm{Var}(\cdot)$ denotes the variance of these input values.

\section{Experiments}

\subsection{Open-vocabulary Aerial Detection Benchmark}
As shown in Fig.~\ref{fig:benchmark}, we evaluate our approach and set the open-vocabulary aerial detection benchmark\footnote{The code
is available at \url{https://github.com/VisionXLab/CastDet}.} on three typical aerial datasets: DIOR~\cite{dior_li2020object}, DOTA~\cite{DOTA8578516} and STAR~\cite{li2024star}. Following the setup of CastDet~\cite{castdet}, we divide the classes into base ($\mathcal{C}_{\text{base}}$) and novel ($\mathcal{C}_{\text{novel}}$) categories. Specifically, we use 16 base classes and 4 novel classes for DIOR, 13 base classes and 2 novel classes for DOTA, and 40 base classes and 8 novel classes for STAR. 
During training, only the annotations for base categories are used. Our experiments cover two tasks: horizontal open-vocabulary detection and oriented open-vocabulary detection.

\subsubsection{Datasets} To establish a robust benchmark, we evaluate our approach using several datasets:

\textbf{DOTA}~\cite{DOTA8578516} comprises 2,806 aerial images from different sensors and platforms with the resolution range from 800 to 4,000, annotated by 15 common categories. We crop the original images into 1,024$\times$1,024 patches with an overlap of 200, and follow the guidelines in MMRotate~\cite{zhou2022mmrotate} to merge the detection results of all patches for evaluation. 

\textbf{DIOR}~\cite{dior_li2020object} provides a large collection of 23,463 images, each with a resolution of $800\times 800$. These images encompass 192,472 instances across 20 object classes. We extract 8,725 images to serve as unlabeled data. Evaluation is conducted on the test set.

\textbf{STAR}~\cite{li2024star} is a large-scale satellite dataset comprising very-high-resolution (VHR) images with sizes ranging from $512 \times 768$ to $27,860 \times 31,096$ pixels. It includes 48 object categories and over 210\textit{k} instances. Following the preprocessing pipeline in MMRotate~\cite{zhou2022mmrotate}, the images are cropped into $1,024 \times 1,024$ patches with an overlap of 200 for training and evaluation.

\textbf{VisDroneZSD}~\cite{VisDrone2023} is a subset of DIOR~\cite{dior_li2020object}, which contains 8,730 images for training and 3,337 images for testing. We follow the generalized zero-shot detection (GZSD) setting in VisDrone2023 Challenge to split the base/novel categories and train/test datasets of DIOR. 

\textbf{NWPU-RESISC45}~\cite{NWPU_RESISC45_7891544} is a large-scale remote sensing classification dataset, covering 45 scene classes with 700 images in each class. NWPU-RESISC45 is used as extra weakly supervised training data for comparison experiment.

\subsubsection{Algorithms} For the open-vocabulary aerial detection task, we establish two types of OVAD benchmarks: HBB-based and OBB-based open-vocabulary detectors. Due to the absence of oriented open-vocabulary detection algorithms, we integrate orientation prediction into several well-established open-vocabulary detection frameworks. We involve significant improvements to these popular frameworks: ViLD~\cite{gu2021open}, GLIP~\cite{9879567glip}, and GroundingDINO~\cite{liu2023groundingdino}. These models are originally designed for horizontal object detection, which are adeptly transformed to predict the orientation of objects, resulting in the development of Oriented ViLD, Oriented GLIP, and Oriented GroundingDINO.

\subsubsection{Evaluation Metrics}
In our evaluation, we adopt mean Average Precision (mAP), mean Average Recall (mAR), and Harmonic Mean (HM) as key metrics. The mAP and mAR metrics are computed at an Intersection over Union (IoU) threshold of 0.5. To provide a comprehensive evaluation, we follow the approach proposed in \cite{VisDrone2023} and incorporate the Harmonic Mean. This metric provides a balanced perspective on overall performance, by evaluating results across base and novel categories. It can be applied to both mAP and mAR metrics, for example, using mAP as an illustration, HM is defined as follows:
\begin{equation}
% \small
% \setlength{\abovedisplayskip}{3pt}
    \mathrm{HM} = 2\frac{{\mathrm{mAP_{base}}}\cdot{\mathrm{mAP_{novel}}}}{{\mathrm{mAP_{base}}}+{\mathrm{mAP_{novel}}}}.
\end{equation}

\subsubsection{Implementation Details}
We implement our method with MMDetection 3.3.0~\cite{mmdetection} and MMRotate 1.0.0~\cite{zhou2022mmrotate} toolbox. For horizontal open-vocabulary detection, we employ Faster R-CNN~\cite{ren2015faster} with ResNet50-C4~\cite{he2016deep} backbone as the base detector. The model is initialized by a pre-trained Soft Teacher~\cite{softteacher_xu2021end} and an external teacher (e.g., RemoteCLIP~\cite{liu2023remoteclip},  SkyCLIP~\cite{wang2024skyscript} or GeoRSCLIP~\cite{zhang2024georsclip}), followed by 10$k$ iterations of training with a batch size of 12. For oriented object open-vocabulary detection, we utilize Oriented R-CNN~\cite{orientedrcnn_Xie_2021_ICCV} as the base detector. The model is initialize with a pre-trained Oriented R-CNN~\cite{orientedrcnn_Xie_2021_ICCV} and the external teacher, followed by 10k iterations of training with a batch size of 8. We adopt Stochastic Gradient Descent (SGD) as the optimizer, using a learning rate of 0.01, with momentum and weight decay parameters set to 0.9 and 0.0001, respectively. To maintain high-quanlity pseudo-labels, we set the RPN foreground threshold and prediction probability threshold $p_0$ to 0.95 and 0.8, respectively, based on the analysis in Fig.~\ref{fig:score_analysis}, to filter out poor predictions.

\subsection{Ablation studies}

\paragraph{Hybrid Training}
To demonstrate the effectiveness of hybrid training in facilitating the discovery of novel categories, we evaluate the RPN recall of the detector across three training paradigms: supervised training (S), closed-vocabulary student-teacher semi-supervised training (S+LT), and our open-vocabulary hybrid training (S+LT+ET).
For fairness, we employ class-agnostic recall, i.e., any object proposed by the RPN is considered successfully detected. As presented in Table \ref{tab:recall_bybrid}, our hybrid training strategy yields substantial improvements in RPN recall for novel categories compared to the semi-supervised approach (e.g., 69.1 vs. 47.7 mAR$_\mathrm{novel}$).

\begin{table*}[h]
% \vspace{-5pt}
  % \setlength{\belowcaptionskip}{-0.05pt}
\caption{Effectiveness of hybrid training in enhancing novel category discovery on VisDroneZSD. A comparative analysis of region proposal recall for different training strategies. S, LT and ET represent the student model, localization teacher model and external teacher model, respectively.} %\Checkmark
% \vspace{-5pt}
\centering
\resizebox{0.78\linewidth}{!}{
\begin{tabular*}{0.9\linewidth}{@{}ccc|cccc|cccc@{}}
\toprule
                   & &      &                                           &                                          &                                         &                              & \multicolumn{4}{c}{novel categories}                                                                                      \\
\multirow{-2}{*}{S} & \multirow{-2}{*}{LT} & \multirow{-2}{*}{ET} & \multirow{-2}{*}{mAR} & \multirow{-2}{*}{mAR$_{\mathrm{base}}$} & \multirow{-2}{*}{mAR$_{\mathrm{novel}}$} & \multirow{-2}{*}{HM}         & airport                      & bball.ct              & track.fld             & windmill                     \\ \hline
\Checkmark & &             & 41.7                                     & 46.0                                    & 24.7                                   & 32.2                & 22.1          & 29.7            & 40.6             & 6.5           \\
\Checkmark & \Checkmark &            & 60.1                                     & \textbf{63.2}                           & 47.7                                   & 54.4                & 31.8          & 80.2            & \textbf{73.0}    & 5.9           \\
\Checkmark & \Checkmark & \Checkmark          & \textbf{63.6}                            & 62.2                                    & \textbf{69.1}                          & \textbf{65.5}       & \textbf{72.1} & \textbf{91.8}   & 71.3             & \textbf{41.3}\\ 
 \bottomrule
\end{tabular*}
}
% \vspace{-5pt}
  \label{tab:recall_bybrid}
\end{table*}

\begin{figure*}[t]
  \centering
   \includegraphics[width=\linewidth]{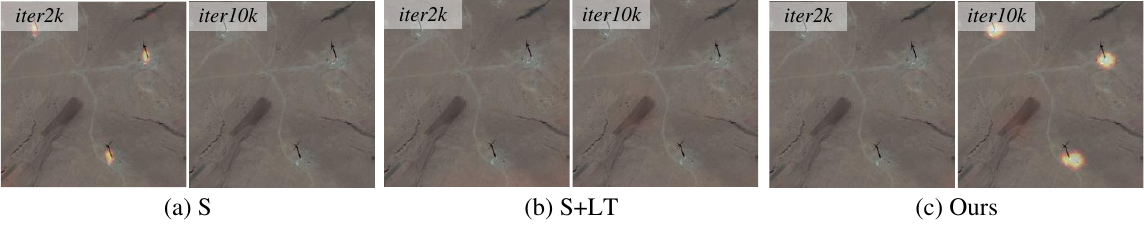}
    \caption{Visualization of RPN foreground confidence during the training process. S, LT and ET denote the student model, localization teacher model and external teacher model, respectively.}
    \label{fig:ab1_score_map}
    % \vspace{-15pt}
\end{figure*}

Fig. \ref{fig:ab1_score_map} illustrates the visualization of RPN foreground confidence score map. The supervised pipeline, while initially capable of discovering novel objects, loses effectiveness in recognizing novel categories as the model focuses on refining classification and bounding box regression of base categories. The semi-supervised pipeline is limited to categories in the labeled dataset and lacks the ability to generalize to novel classes. In contrast, our method progressively enhances detection performance for novel categories, demonstrating a robust capability for open-vocabulary aerial detection.

\begin{figure*}[t]
  \centering
\includegraphics[width=\linewidth]{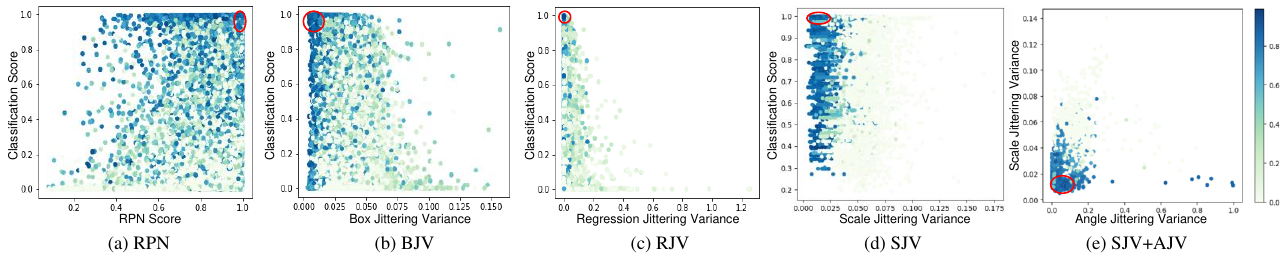}
\caption{Visualization of three types of box selection strategies. The figures shows the correlation among IoU, classification score, and (a) RPN score, (b) box-jittering variance, (c) regression-jittering variance, (d) oriented box scale-jittering variance, and (e) oriented angle-jitter variance respectively. Among them, IoU is represented by the color bar.}
   \label{fig:score_analysis}
\end{figure*}

\begin{figure*}[t]
  \centering
\includegraphics[width=0.95\linewidth]{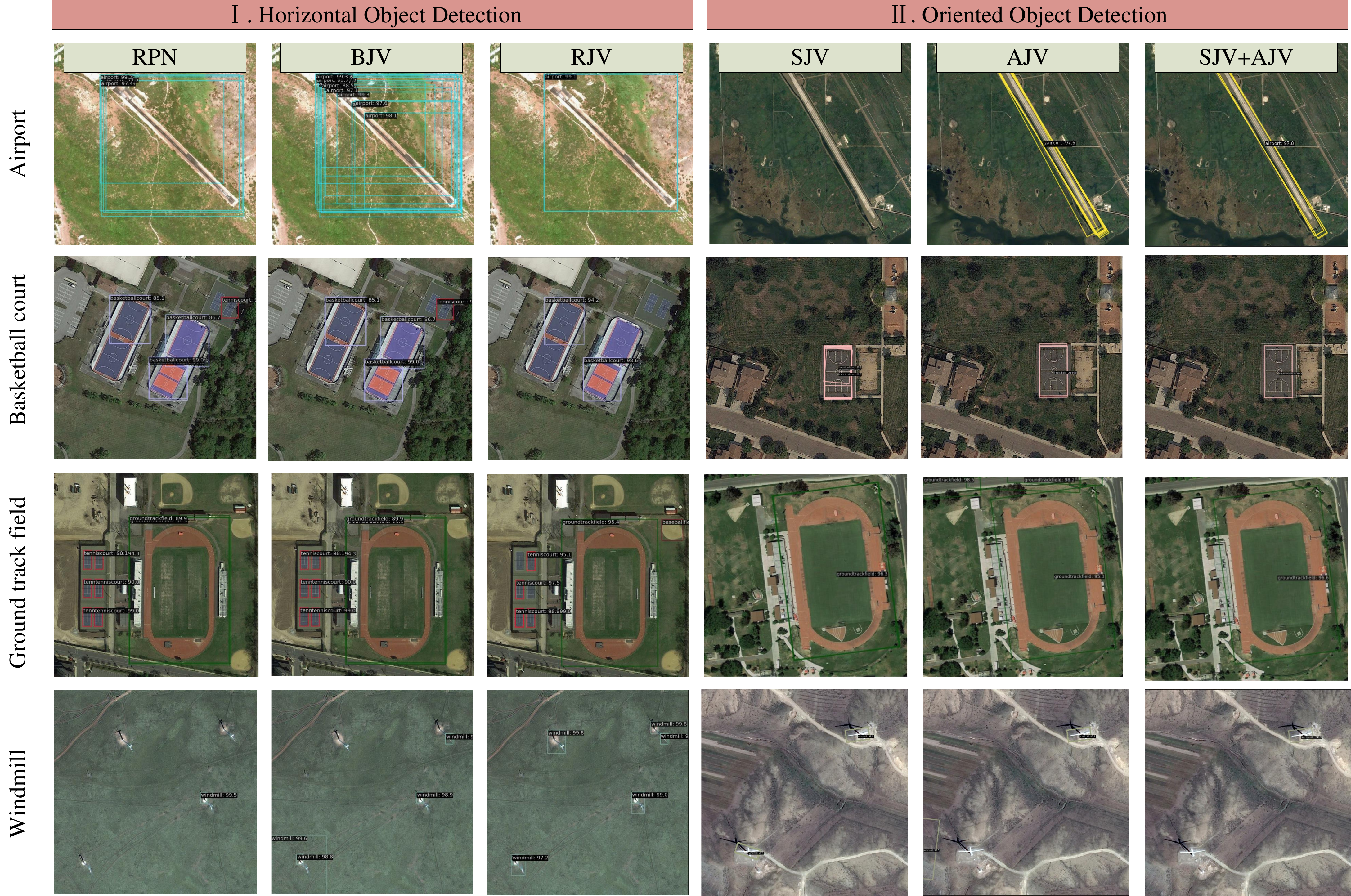}
   \caption{Visualization of pseudo-labels of different box selection strategies on DIOR dataset. Left: horizontal pseudo-boxes selected by RPN, BJV and RJV score. Right: oriented pseudo-boxes selected by AJV and BJV score.}
   \label{fig:box_selection_vis}
\end{figure*}

\paragraph{H-Box Selection Strategy}
We evaluate the effectiveness of different horizontal box selection metrics, including the RPN Score, RJV, and BJV, with the results summarized in Table~\ref{tab:ab7_box_selection}. The regression jittering strategy achieves a superior performance of 47.8\% mAP$_\mathrm{novel}$. As shown in Fig.~\ref{fig:score_analysis}(a)$\sim$(c), regression jittering exhibits a stronger correlation with both the classification score and the IoU score, enabling the selection of more accurate pseudo-labels, which enhances the training process. The qualitative results of pseudo-labels under different box selection strategies are presented in Fig.~\ref{fig:box_selection_vis}.

\begin{table}[h]
\caption{Comparison on horizontal box selection strategies on VisDroneZSD.} %\Checkmark
\centering
\begin{tabular}{c|cccc}
\toprule
Strategy      & mAP & mAP$_{\mathrm{base}}$ & mAP$_{\mathrm{novel}}$ & HM            \\ \hline
RPN Score     & 39.5          & 38.6            & 43.3             & 40.8          \\
BJV Score & 39.2          & 38.0            & 43.6             & 40.6          \\
RJV Score& \textbf{40.7} & \textbf{39.0}   & \textbf{47.8}    & \textbf{42.9} \\ \bottomrule
\end{tabular}
  \label{tab:ab7_box_selection}
\end{table}

\paragraph{R-Box Selection Strategy}
We compare the effects of two oriented box selection strategies: SJV and AJV. Fig.\ref{fig:score_analysis}(d)$\sim$(e) illustrates improved score correlation when employing SJV and AJV, which enhances the model training. 
Furthermore, Table~\ref{tab:ab_rbox_selection} shows that using both AJV and SJV score yields better results, e.g., achieving 24.3\% mAP$_{\mathrm{novel}}$ on DIOR-R and 34.9\% mAP$_{\mathrm{novel}}$ on DOTA-R.
The visualization of selected pseudo-boxes is shown in Fig.~\ref{fig:box_selection_vis}.

\begin{table}[h]
\caption{Comparison on oriented box selection strategies.}
\centering
\begin{tabular}{c|cccc|cccc}
\toprule
\multirow{2}{*}{Strategy}
& \multicolumn{4}{c|}{DIOR}
& \multicolumn{4}{c}{DOTA} \\
\cline{3-4} \cline{7-8}
& mAP & mAP$_{\mathrm{base}}$ & mAP$_{\mathrm{novel}}$ & HM
& mAP & mAP$_{\mathrm{base}}$ & mAP$_{\mathrm{novel}}$ & HM \\
\hline
RPN Score
& 45.7 & 51.1 & 23.9 & 32.6
& 57.6 & \textbf{62.2} & 27.6 & 38.3 \\
AJV Score
& 45.7 & 51.1 & 24.2 & 32.8
& 57.4 & 61.6 & 30.2 & 40.6 \\
SJV Score
& 45.9 & \textbf{51.4} & 23.9 & 32.6
& 57.2 & 61.1 & 31.6 & 41.7 \\
AJV+SJV
& \textbf{45.9} & 51.3 & \textbf{24.3} & \textbf{33.0}
& \textbf{58.1} & 61.7 & \textbf{34.9} & \textbf{44.6} \\
\bottomrule
\end{tabular}
\label{tab:ab_rbox_selection}
\end{table}

\paragraph{Dynamic Label Queue}
To demonstrate the effectiveness of dynamic label queue, we conduct experiments on whether to adopt a dynamic update strategy or whether to use a label queue for storing pseudo-labels. Specifically, we consider the following settings: \rmnum{1}) training the model without a dynamic queue; \rmnum{2}) incorporating teacher models to generate pseudo-labels at each iteration, and directly using these labels to train the student model without storing them in a queue; \rmnum{3}) constructing a static pseudo queue; and \rmnum{4}) implementing the proposed dynamic label queue. As shown in Table~\ref{tab:ab2_dynamic_queue}, adopting either a dynamic update strategy or a label queue leads to substantial performance improvements, with the proposed dynamic label queue achieving the most significant gains.

\begin{figure*}[t]
  \centering
\includegraphics[width=\linewidth]{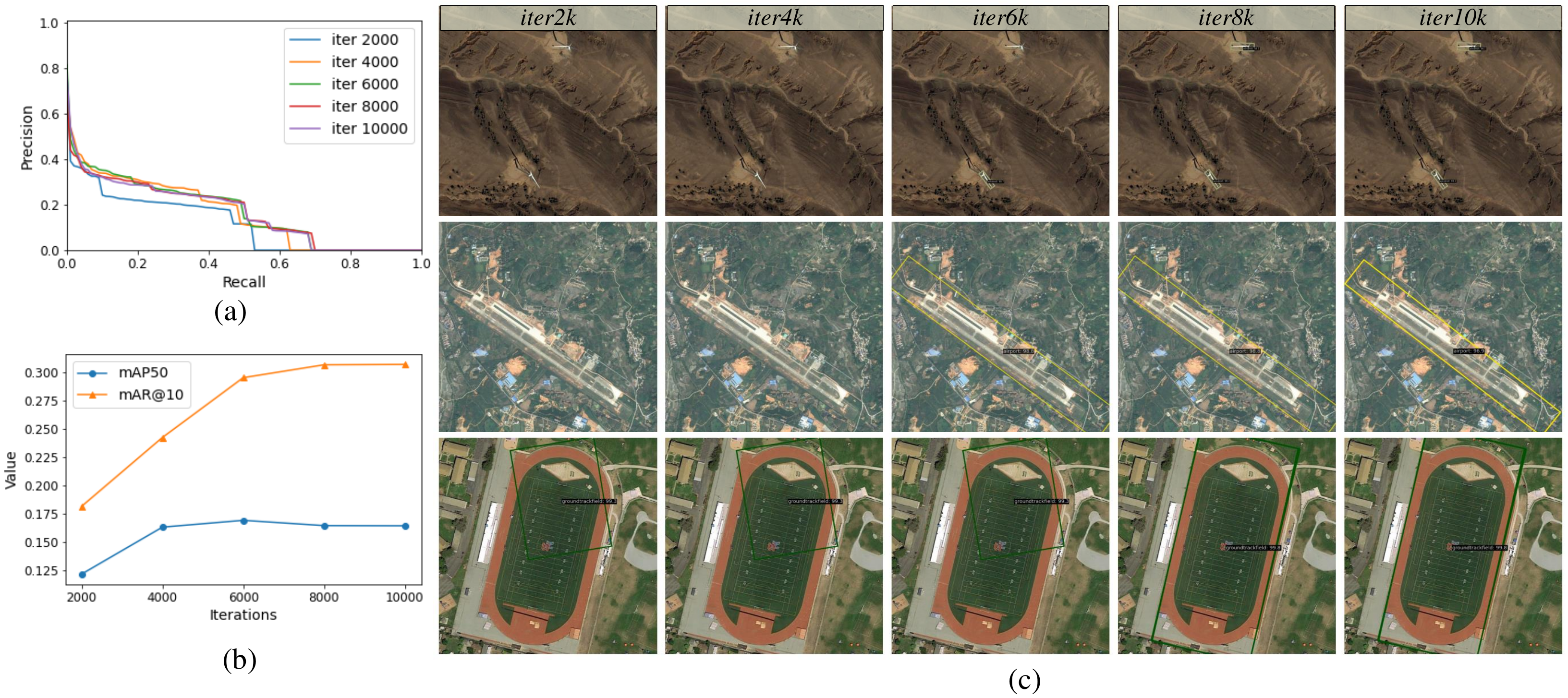}
   \caption{Statistics and visualizations for novel categories of pseudo-labels in dynamic label queue. (a) Precision-recall curve for different iterations. (b) The mAP and mAR curve on iterations. (c) Visualization of pseudo-labels across different iterations.}
   \label{fig:queue_iter_analysis}
\end{figure*}

To further illustrate that the dynamic queue can obtain more accurate pseudo-labels, we analyze the pseudo-labels in the dynamic label queue across different training iterations.
As illustrated in Fig.~\ref{fig:queue_iter_analysis}, with the model iterations, the precision-recall (PR) curve expands outward (Fig.~\ref{fig:queue_iter_analysis} (a)), and mAP$_{50}^{\mathrm{novel}}$  of pseudo-labels shows significant improvements (Fig.~\ref{fig:queue_iter_analysis} (b)).
This indicates that the dynamic label queue maintains richer and more accurate pseudo-labels as training progresses.

Moreover,  Fig.~\ref{fig:queue_iter_analysis}(b) shows that the quality of pseudo-labels saturates at an mAP$_{50}^{\mathrm{novel}}$ around 16.9\% after 4$k$ iterations. This is expected, as a simple combination of the detector with RemoteCLIP is suboptimal for open-vocabulary detection. However, we can improve the mAP$_{50}^{\mathrm{novel}}$ to 43.3\% by employing our hybrid training mechanism. Fig.~\ref{fig:queue_iter_analysis}(c) visualizes the pseudo-labels across different iterations. We can see that the model generates more accurate pseudo-labels (last row) and demonstrates an improved ability to discover novel objects (first two rows) as the training goes.

\begin{table}[h]
\caption{Ablation on dynamic label queue on VisDroneZSD.} %\Checkmark
\centering
\begin{tabular}{c|cc|cccc}
\toprule
ID & Dynamic                   & Queue                     & mAP & mAP$_{\mathrm{base}}$ & mAP$_{\mathrm{novel}}$ & HM   \\ \hline
 \rmnum{1} &                         &                           & 11.6  & 9.7                     & 19.3                     & 12.9 \\
\rmnum{2} & \Checkmark &                           & 37.7 & 36.3 & \textbf{43.5} & 39.6 \\
  \rmnum{3} &                        & \Checkmark & 38.1 & 37.2 & 41.5 & 39.2 \\
\rmnum{4} & \Checkmark & \Checkmark & \textbf{39.5} & \textbf{38.6} & 43.3 & \textbf{40.8} 
\\ \bottomrule
\end{tabular}
% \vspace{-5pt}
\label{tab:ab2_dynamic_queue}
\end{table}

\paragraph{Cross Evaluation}
Some popular detectors for natural images, such as GroundingDINO~\cite{liu2023groundingdino} and GLIP~\cite{9879567glip}, underperform on novel aerial categories due to a lack of a robust strategy for discovering novel aerial objects. Thus, we provide them with pseudo-labels filtered by our box selection strategies, i.e., AJV and RJV. As shown in Table~\ref{tab:ab_gdino_labels}, this leads to significant improvements, e.g., 6.6\% mAP$_{\mathrm{novel}}$ gains for GLIP and 11.8\% mAP$_{\mathrm{novel}}$ gains for GroundingDINO, highlighting the effectiveness of our approach in bridging the gap between general object detection frameworks and the specialized requirements of aerial imagery.

\begin{table}[h]
% % \renewcommand{\arraystretch}{1.2}
% \setlength\tabcolsep{3pt} % set the column space
\caption{Boost GLIP/GroundingDINO with our pseudo-labeling strategies on DOTA. $\ddagger$: Use pseudo-labels generated by our pseudo-labeling strategies to train the model.}
\centering
\begin{tabular}{c|cccc}
\toprule
Method        & mAP  & mAP$_{\mathrm{base}}$ & mAP$_{\mathrm{novel}}$ & HM   \\ \hline
GLIP          & 57.5 & 65.2                  & 8.0                      & 14.2 \\
GroundingDINO & 58.0   & 66.0                    & 5.8                    & 10.6 \\ \rowcolor[HTML]{EFEFEF} 
GLIP $^\ddagger$         & 59.7 & 66.6                  & 14.6                   & 24.0   \\ \rowcolor[HTML]{EFEFEF} 
GroundingDINO $^\ddagger$ & 47.0   & 51.5                  & 17.6                   & 26.3 \\ \bottomrule
\end{tabular}
% \vspace{-5pt}
\label{tab:ab_gdino_labels}
\end{table}

\paragraph{External Teacher} We evaluate different external teachers by varying both their backbone architectures (e.g., R50 vs. ViT-L-14) and vision–language models (e.g., RemoteCLIP~\cite{liu2023remoteclip}, GeoRSCLIP~\cite{zhang2024georsclip}, and SkyCLIP~\cite{wang2024skyscript}). As shown in Table~\ref{tab:ab_external_teacher}, CastDet maintains strong overall performance across different choices of external teachers, with stronger architectures (e.g., SkyCLIP and ViT-L-14 backbone) achieving better performance on novel categories.

\begin{table*}[h]
\caption{Ablation on external teachers.}
% \vspace{-5pt}
\centering
\resizebox{0.8\linewidth}{!}{
\begin{tabular*}{0.82\linewidth}{c|cc|cccc}
\toprule
Dataset                      & External Teacher & Backbone & mAP           & mAP$_{\mathrm{base}}$ & mAP$_{\mathrm{novel}}$ & HM            \\ \hline
\multirow{5}{*}{VisDroneZSD} & RemoteCLIP~\cite{liu2023remoteclip}       & R50     & 40.5          & 39.0                  & 46.3                   & 42.3          \\
                             & RemoteCLIP~\cite{liu2023remoteclip}       & VIT-L-14 & 40.8          & 38.2                  & \textbf{51.2}          & 43.8          \\
                             & GeoRSCLIP~\cite{zhang2024georsclip}        & VIT-B-32 & \textbf{41.8} & \textbf{40.8}         & 45.6                   & 43.1          \\
                             & GeoRSCLIP~\cite{zhang2024georsclip}        & VIT-L-14 & 41.1          & 39.4                  & 48.2                   & 43.3          \\
                             & SkyCLIP~\cite{wang2024skyscript}          & VIT-L-14 & 41.2          & 38.8                  & 51.0                   & \textbf{44.0} \\ \hline
\multirow{5}{*}{DIOR}        & RemoteCLIP~\cite{liu2023remoteclip}       & R50     & 61.6          & 65.7                  & 45.2                   & 53.6          \\
                             & RemoteCLIP~\cite{liu2023remoteclip}       & VIT-L-14 & 61.6          & 64.4                  & \textbf{50.5}          & \textbf{56.6} \\
                             & GeoRSCLIP~\cite{zhang2024georsclip}        & VIT-B-32 & \textbf{62.4} & \textbf{66.6}         & 45.2                   & 53.9          \\
                             & GeoRSCLIP~\cite{zhang2024georsclip}        & VIT-L-14 & 61.6          & 65.1                  & 47.6                   & 55.0          \\
                             & SkyCLIP~\cite{wang2024skyscript}          & VIT-L-14 & 61.6          & 64.5                  & 50.1                   & 56.4  \\ \bottomrule  
\end{tabular*}
}
% \vspace{-5pt}
  \label{tab:ab_external_teacher}
\end{table*}

\paragraph{Label Fraction Experiments} To demonstrate that our model is well suited for the aerial scenario where labeled data is limited, we conduct label fraction experiments with 34\%, 50\% and full labeled data.
Table~\ref{tab:ab5_label_fraction} shows that even with only 34\% of the original labeled data, the performance remains comparable, with only a minor reduction of 0.9\% mAP.

\begin{table}[h]
\caption{Ablation on fraction of labeled data on VisDroneZSD.}\label{tab:ab5_label_fraction}
\centering
\begin{tabular}{c|cccc}
\toprule
Label Fraction & mAP & mAP$_{\mathrm{base}}$ & mAP$_{\mathrm{novel}}$ & HM             \\ \hline
34\%           & 38.6          & 38.0            & 41.0             & 39.5          \\
50\%           & 38.8          & 37.7            & \textbf{43.4}    & 40.4          \\
100\%          & \textbf{39.5} & \textbf{38.6}   & 43.3             & \textbf{40.8} \\ \bottomrule
\end{tabular}
\end{table}

\paragraph{Loss weights of $\mathcal{L}_s$, $\mathcal{L}_u$ and $\mathcal{L}_d$.} Table~\ref{tab:ab_loss_weights} presents the performance comparison with different loss weight configurations.
We observe that setting $\alpha=1.0$, $\beta=2.0$ and $\gamma=1.0$ could be better for novel categories.

\begin{table}[h]
\caption{Ablation with different loss weights defined in Equ.~(\ref{equ:losses}) on DOTA.}
\centering
\begin{tabular}{ccc|cccc}
\toprule
$\alpha$ & $\beta$ & $\gamma$ & mAP  & mAP$_{\mathrm{base}}$ & mAP$_{\mathrm{novel}}$ & HM   \\ \hline
1.0      & 1.0     & 1.0      & \textbf{62.6} & \textbf{67.7}                  & 29.6                   & 41.2 \\
2.0      & 1.0     & 1.0      & 59.1 & 64.2                  & 26.0                   & 37.1 \\
1.0      & 2.0     & 1.0      & 58.1 & 61.7                  & \textbf{34.9}                   & \textbf{44.6} \\
1.0      & 1.0     & 2.0      & 60.0 & 65.0                  & 27.7                   & 38.9 \\ \bottomrule
\end{tabular}
\label{tab:ab_loss_weights}
\end{table}

\paragraph{Background Embeddings}
% Mean, zero, learnable.
We compare three common background embeddings: fixed all-zero embedding~\cite{zhong2022regionclip}, fixed mean embedding~\cite{fixMean_rahman2018zero}, and learnable background embedding~\cite{learnBG_zheng2020background}. As shown in Table~\ref{tab:ab_bg_embeddings}, the choice of background embedding has slight impact on our experimental results, with the learnable embedding showing better performance.

\begin{table}[h]
\caption{Ablation with Different Background Embeddings on DOTA.}
\centering
\begin{tabular}{c|cccc}
\toprule
BG Embedding & mAP           & mAP$_{\mathrm{base}}$ & mAP$_{\mathrm{novel}}$ & HM            \\ \hline
Zero         & \textbf{58.5} & \textbf{62.3}         & 33.9                   & 43.9          \\
Normalize    & 58.4          & 62.1                  & 34.2                   & 44.2          \\
Learnable    & 58.1          & 61.7                  & \textbf{34.9}          & \textbf{44.6}\\ \bottomrule
\end{tabular}
\label{tab:ab_bg_embeddings}
\end{table}

\paragraph{Prompt Templates.}
We evaluate four types of prompt templates, including T1: ``[\underline{category}]'', T2: ``a [\underline{category}]'', T3: ``a satellite photo of [\underline{category}]'', and T4: ``a photo of [\underline{category}]''. According to Table~\ref{tab:ab_prompt_templates}, T4 is more suitable.

\begin{table}[h]
\caption{Ablation with different prompt templates on DOTA.}
\centering
\begin{tabular}{c|cccc}
\toprule
\multicolumn{1}{c|}{Template}                                                               & mAP           & mAP$_{\mathrm{base}}$ & mAP$_{\mathrm{novel}}$ & HM            \\ \hline
T1 & 56.7 & \textbf{61.8}         & 24.0& 34.6          \\
T2 & 55.9          & 60.3                  & 27.3                   & 37.6          \\
T3 & 57.5          & 61.3                  & 32.7          & 42.7 \\
T4 & \textbf{58.1} & 61.7                  & \textbf{34.9}          & \textbf{44.6}\\ \bottomrule
\end{tabular}
\label{tab:ab_prompt_templates}
\end{table}

\paragraph{Function $\mathrm{f}(\cdot)$ of AJV Score }
% identity, sin()
Table~\ref{tab:ab_ajv_funcf} compares two types of functions $\mathrm{f}(\cdot)$ in Equ.(\ref{equ:ajv}), i.e., $\mathrm{Identity}(\cdot)$ and $\sin (\cdot)$. The results indicate that using $\sin (\cdot)$ function for angle transformation yields better performance. This improvement is attributed to the fact that the oriented box representation exhibits discontinuities at boundaries, and the $\sin (\cdot)$ function helps to smooth these discontinuities, resulting in more reliable variance estimates.

\begin{table}[h]
\caption{Ablation on Function $\mathrm{f}(\cdot)$ of AJV Score on DOTA.}
\centering
\begin{tabular}{c|cccc}
\toprule
$\mathrm{f}(\cdot)$& mAP  & mAP$_{\mathrm{base}}$ & mAP$_{\mathrm{novel}}$ & HM   \\\hline
$\mathrm{Identity}(\cdot)$ & 57.4 & 61.6                  & 30.2                   & 40.6 \\
$\sin (\cdot)$             & \textbf{59.1} & \textbf{63.3}                  & \textbf{31.2}                   & \textbf{41.8} \\ \bottomrule
\end{tabular}
\label{tab:ab_ajv_funcf}
\end{table}

\paragraph{Computation Costs.}
Table~\ref{tab:ab_compute} compares the computational costs of various methods in terms of FLOPs, parameter count, memory consumption, and inference speed (FPS), measured on an NVIDIA A6000 GPU. During training, CastDet utilizes teacher models to guide the student model training. After training, both the teacher and the student models are capable of detecting objects independently. Therefore, CastDet demonstrates superior efficiency, with the lowest parameter count (42.17M) and the highest FPS (9.8).

\begin{table}[h]
\caption{Comparison of computational costs across different methods.}
\centering
\begin{tabular}{c|cccc}
\toprule
Method          & Flops(G)$\downarrow$ & Params(M)$\downarrow$ & Mem(GB)$\downarrow$  & FPS$\uparrow$ \\ \hline
ViLD           & {\ul 150.38}  & {\ul 80.49}   & \textbf{313} & {\ul 8.7}\\
GLIP           & \textbf{89.66}    & 147.70  & 568 & 5.4\\
GroundingDINO & 243.22  & 172.15  & 662 & 6.2\\
CastDet      & {\ul 150.38}  & \textbf{42.17} &  {\ul 332} & \textbf{9.8}\\ \bottomrule
\end{tabular}
% \vspace{-5pt}
\label{tab:ab_compute}
\end{table}

\subsection{Comparisons with state-of-the-art methods}

\begin{table*}[!ht]
\caption{Evaluation on CLIPs (e.g., RemoteCLIP, CLIP) with ground-truth (GT) or RPN proposals.} %\Checkmark
% \vspace{-5pt}
\centering
\resizebox{0.85\linewidth}{!}{
\begin{tabular*}{0.92\linewidth}{@{}cc|cccc|cccc@{}}
\toprule
       &          & \multicolumn{4}{c|}{VisDroneZSD}                                                  & \multicolumn{4}{c}{COCO}                                                       \\ \hline
Method & Proposal & mAP           & mAP$_{\mathrm{base}}$ & mAP$_{\mathrm{novel}}$ & HM            & mAP           & mAP$_{\mathrm{base}}$ & mAP$_{\mathrm{novel}}$ & HM            \\ \hline
CLIP   & GT       & 37.6          & 30.4                  & \textbf{66.4}          & 41.7          & 39.8          & 39.1                  & \textbf{41.5}          & 40.3          \\
Ours   & GT       & \textbf{47.3} & \textbf{46.6}         & 49.9                   & \textbf{48.2} & \textbf{51.8} & \textbf{55.5}         & 41.4                   & \textbf{47.4} \\ \hline
CLIP   & RPN      & 11.6          & 9.7                   & 19.3                   & 12.9          & 14.7          & 15.2                  & 13.2                   & 14.2          \\
Ours   & RPN      & \textbf{38.1} & \textbf{36.6}         & \textbf{44.2}          & \textbf{40.0} & \textbf{37.0} & \textbf{40.6}         & \textbf{27.1}          & \textbf{32.5} \\ \bottomrule
\end{tabular*}
}
% \vspace{-5pt}
\label{tab:com2clip}
\end{table*}

\begin{table*}[!ht]
\renewcommand{\arraystretch}{1.2}
\caption{Horizontal open-vocabulary detection benchmark on DIOR dataset.  $\mathcal{T}_{\mathrm{novel}}$ indicates whether the novel classes need to be pre-known.  $\mathcal{I}_{\mathrm{cls}}$ denotes whether extra classification or caption datasets are required during training.  $\dagger$: The best result on the zero-shot object detection leaderboard of VisDrone2023. $\ddagger$: The results of our own implementation. The short names for categories are defined as: Airport (APO), Basketball court (BC), Ground track field (GTF) and Windmill (WM). S and LT denote the student model and the localization teacher model respectively.}
% \vspace{-5pt}
\centering
\resizebox{1.\linewidth}{!}{
\begin{tabular*}{1.45\linewidth}{@{}c|c|cccc|cccc|cccc@{}}
\toprule
Dataset & Method & Detector & Backbone & $\mathcal{T}_{\mathrm{novel}}$ & $\mathcal{I}_{\mathrm{cls}}$ & APO & BC & GTF & WM & mAP & mAP$_{\mathrm{base}}$ & mAP$_{\mathrm{novel}}$ & HM \\ \hline
 & MultiModel & - & - & \xmark & \xmark & - & - & - & - & - & - & - & 26.7$^\dagger$ \\
 & ViLD$^\ddagger$~\cite{gu2021open} & Faster R-CNN & R50 & \cmark & \xmark & 9.1 & 11.8 & 26.9 & 9.1 & 25.6 & 28.5 & 14.2 & 19.0 \\
 & OV-DETR~\cite{zang2022open} & Deformable DETR & R50 & \cmark & \xmark & 29.9 & 20.5 & 26.8 & 5.2 & 28.7 & 30.8 & 20.6 & 24.7 \\
 & Detic~\cite{zhou2022detic} & CenterNet2 & R50 & \xmark & \cmark & 2.0 & 2.2 & 15.0 & 0.0 & 16.8 & 19.8 & 4.8 & 7.7 \\
 & GroundingDINO~\cite{liu2023groundingdino} & GroundingDINO & R50 & \xmark & \cmark & 0.5 & 3.4 & 9.1 & 0.2 & 33.0 & {\ul 40.5} & 3.3 & 6.1 \\
 & GLIP~\cite{9879567glip} & GLIP & R50 & \xmark & \cmark &  0.1 & 5.7 & 14.9 & 0.0 & 28.3 & 34.9 & 5.2 & 9.0 \\
 & GLIP~\cite{9879567glip} & GLIP & Swin-T & \xmark & \cmark & 0.0 & 3.3 & 9.1 & 9.1 & 33.8 & \textbf{41.0} & 5.4 & 9.5 \\
 & BARON~\cite{wu2023baron} & Faster R-CNN & R50 & \xmark & \cmark & 22.8 & 18.3 & 34.0 & 3.0 & 27.4 & 29.4 & 19.5 & 23.5 \\
 & YOLO-World~\cite{Cheng2024YOLOWorld} & YOLOv8-M & YOLOv8-M & \xmark & \cmark & 0.0 & 20.1 & 13.7 & 0.0 & 32.9 & 39.1 & 8.5 & 13.9 \\
 & \cellcolor[HTML]{EFEFEF}CastDet (\textit{S}) & \cellcolor[HTML]{EFEFEF}Faster R-CNN & \cellcolor[HTML]{EFEFEF}R50 & \cellcolor[HTML]{EFEFEF}\xmark & \cellcolor[HTML]{EFEFEF}\xmark & \cellcolor[HTML]{EFEFEF}\textbf{57.3} & \cellcolor[HTML]{EFEFEF}75.6 & \cellcolor[HTML]{EFEFEF}33.3 & \cellcolor[HTML]{EFEFEF}\textbf{23.2} & \cellcolor[HTML]{EFEFEF}36.4 & \cellcolor[HTML]{EFEFEF}33.7 & \cellcolor[HTML]{EFEFEF}\textbf{47.4} & \cellcolor[HTML]{EFEFEF}39.4 \\
 & \cellcolor[HTML]{EFEFEF}CastDet (\textit{LT}) & \cellcolor[HTML]{EFEFEF}Faster R-CNN & \cellcolor[HTML]{EFEFEF}R50 & \cellcolor[HTML]{EFEFEF}\xmark & \cellcolor[HTML]{EFEFEF}\xmark & \cellcolor[HTML]{EFEFEF}{\ul 50.4} & \cellcolor[HTML]{EFEFEF}{\ul 77.9} & \cellcolor[HTML]{EFEFEF}{\ul 37.4} & \cellcolor[HTML]{EFEFEF}{\ul 20.1} & \cellcolor[HTML]{EFEFEF}{\ul 38.5} & \cellcolor[HTML]{EFEFEF}36.5 & \cellcolor[HTML]{EFEFEF}{\ul 46.5} & \cellcolor[HTML]{EFEFEF}{\ul 40.9} \\
 & \cellcolor[HTML]{EFEFEF}CastDet (\textit{S}) & \cellcolor[HTML]{EFEFEF}Faster R-CNN & \cellcolor[HTML]{EFEFEF}R50 & \cellcolor[HTML]{EFEFEF}\cmark & \cellcolor[HTML]{EFEFEF}\xmark & \cellcolor[HTML]{EFEFEF}37.3 & \cellcolor[HTML]{EFEFEF}77.6 & \cellcolor[HTML]{EFEFEF}34.6 & \cellcolor[HTML]{EFEFEF}16.3 & \cellcolor[HTML]{EFEFEF}35.2 & \cellcolor[HTML]{EFEFEF}33.6 & \cellcolor[HTML]{EFEFEF}41.5 & \cellcolor[HTML]{EFEFEF}37.1 \\
\multirow{-12}{*}{VisDroneZSD} & \cellcolor[HTML]{EFEFEF}CastDet (\textit{LT}) & \cellcolor[HTML]{EFEFEF}Faster R-CNN & \cellcolor[HTML]{EFEFEF}R50 & \cellcolor[HTML]{EFEFEF}\cmark & \cellcolor[HTML]{EFEFEF}\xmark & \cellcolor[HTML]{EFEFEF}47.5 & \cellcolor[HTML]{EFEFEF}\textbf{78.3} & \cellcolor[HTML]{EFEFEF}\textbf{43.3} & \cellcolor[HTML]{EFEFEF}16.0 & \cellcolor[HTML]{EFEFEF}\textbf{40.5} & \cellcolor[HTML]{EFEFEF}39.0 & \cellcolor[HTML]{EFEFEF}46.3 & \cellcolor[HTML]{EFEFEF}\textbf{42.3} \\
\hline
 & ViLD$^\ddagger$~\cite{gu2021open} & Faster R-CNN & R50 & \cmark & \xmark & 9.1 & 11.8 & 26.7 & 9.1 & 45.7 & 53.5 & 14.2 & 22.4 \\
 & OV-DETR~\cite{zang2022open} & Deformable DETR & R50 & \cmark & \xmark & 29.1 & 19.7 & 25.6 & 5.2 & 54.0 & 62.6 & 19.9 & 30.2 \\
 & Detic~\cite{zhou2022detic} & CenterNet2 & R50 & \xmark & \cmark & 0.7 & 2.2 & 11.2 & 0.0 & 36.9 & 45.3 & 3.5 & 6.5 \\
 & GroundingDINO~\cite{liu2023groundingdino} & GroundingDINO & R50 & \xmark & \cmark & 0.3 & 3.3 & 9.1 & 0.2 & 57.3 & \textbf{70.8} & 3.2 & 6.2 \\
 & GLIP~\cite{9879567glip} & GLIP & R50 & \xmark & \cmark &  0.0 & 2.1 & 13.4 & 0.0 & 50.3 & 63.6 & 3.9 & 7.3 \\
 & GLIP~\cite{9879567glip} & GLIP & Swin-T & \xmark & \cmark & 0.0 & 2.4 & 9.1 & 3.0 & 56.0 & 69.1 & 3.6 & 6.9 \\
 & BARON~\cite{wu2023baron} & Faster R-CNN & R50 & \xmark & \cmark & 14.1 & 16.6 & 30.4 & 0.1 & 50.9 & 59.8 & 15.3 & 24.4 \\
 & YOLO-World~\cite{Cheng2024YOLOWorld} & YOLOv8-M & YOLOv8-M & \xmark & \cmark & 0.0 & 19.8 & 12.2 & 0.0 & 57.7 & {\ul 70.2} & 8.0 & 14.4 \\
 & \cellcolor[HTML]{EFEFEF}CastDet (\textit{S}) & \cellcolor[HTML]{EFEFEF}Faster R-CNN & \cellcolor[HTML]{EFEFEF}R50 & \cellcolor[HTML]{EFEFEF}\xmark & \cellcolor[HTML]{EFEFEF}\xmark & \cellcolor[HTML]{EFEFEF}\textbf{55.9} & \cellcolor[HTML]{EFEFEF}75.4 & \cellcolor[HTML]{EFEFEF}32.7 & \cellcolor[HTML]{EFEFEF}\textbf{22.7} & \cellcolor[HTML]{EFEFEF}56.5 & \cellcolor[HTML]{EFEFEF}58.9 & \cellcolor[HTML]{EFEFEF}\textbf{46.7} & \cellcolor[HTML]{EFEFEF}52.1 \\
 & \cellcolor[HTML]{EFEFEF}CastDet (\textit{LT}) & \cellcolor[HTML]{EFEFEF}Faster R-CNN & \cellcolor[HTML]{EFEFEF}R50 & \cellcolor[HTML]{EFEFEF}\xmark & \cellcolor[HTML]{EFEFEF}\xmark & \cellcolor[HTML]{EFEFEF}{\ul 47.1} & \cellcolor[HTML]{EFEFEF}76.8 & \cellcolor[HTML]{EFEFEF}{\ul 35.9} & \cellcolor[HTML]{EFEFEF}{\ul 20.0} & \cellcolor[HTML]{EFEFEF}{\ul 59.3} & \cellcolor[HTML]{EFEFEF}62.8 & \cellcolor[HTML]{EFEFEF}45.0 & \cellcolor[HTML]{EFEFEF}{\ul 52.4} \\
 & \cellcolor[HTML]{EFEFEF}CastDet (\textit{S}) & \cellcolor[HTML]{EFEFEF}Faster R-CNN & \cellcolor[HTML]{EFEFEF}R50 & \cellcolor[HTML]{EFEFEF}\cmark & \cellcolor[HTML]{EFEFEF}\xmark & \cellcolor[HTML]{EFEFEF}34.0 & \cellcolor[HTML]{EFEFEF}{\ul 77.4} & \cellcolor[HTML]{EFEFEF}33.5 & \cellcolor[HTML]{EFEFEF}16.3 & \cellcolor[HTML]{EFEFEF}56.9 & \cellcolor[HTML]{EFEFEF}61.0 & \cellcolor[HTML]{EFEFEF}40.3 & \cellcolor[HTML]{EFEFEF}48.5 \\
\multirow{-11}{*}{DIOR} & \cellcolor[HTML]{EFEFEF}CastDet (\textit{LT}) & \cellcolor[HTML]{EFEFEF}Faster R-CNN & \cellcolor[HTML]{EFEFEF}R50 & \cellcolor[HTML]{EFEFEF}\cmark & \cellcolor[HTML]{EFEFEF}\xmark & \cellcolor[HTML]{EFEFEF}45.4 & \cellcolor[HTML]{EFEFEF}\textbf{78.2} & \cellcolor[HTML]{EFEFEF}\textbf{41.3} & \cellcolor[HTML]{EFEFEF}15.9 & \cellcolor[HTML]{EFEFEF}\textbf{61.6} & \cellcolor[HTML]{EFEFEF}65.7 & \cellcolor[HTML]{EFEFEF}{\ul 45.2} & \cellcolor[HTML]{EFEFEF}\textbf{53.6} \\ \bottomrule
\end{tabular*}
}
  \label{tab:com2best}
  % \vspace{-5pt}
\end{table*}

\begin{figure*}[!ht]
  \centering
\includegraphics[width=\linewidth]{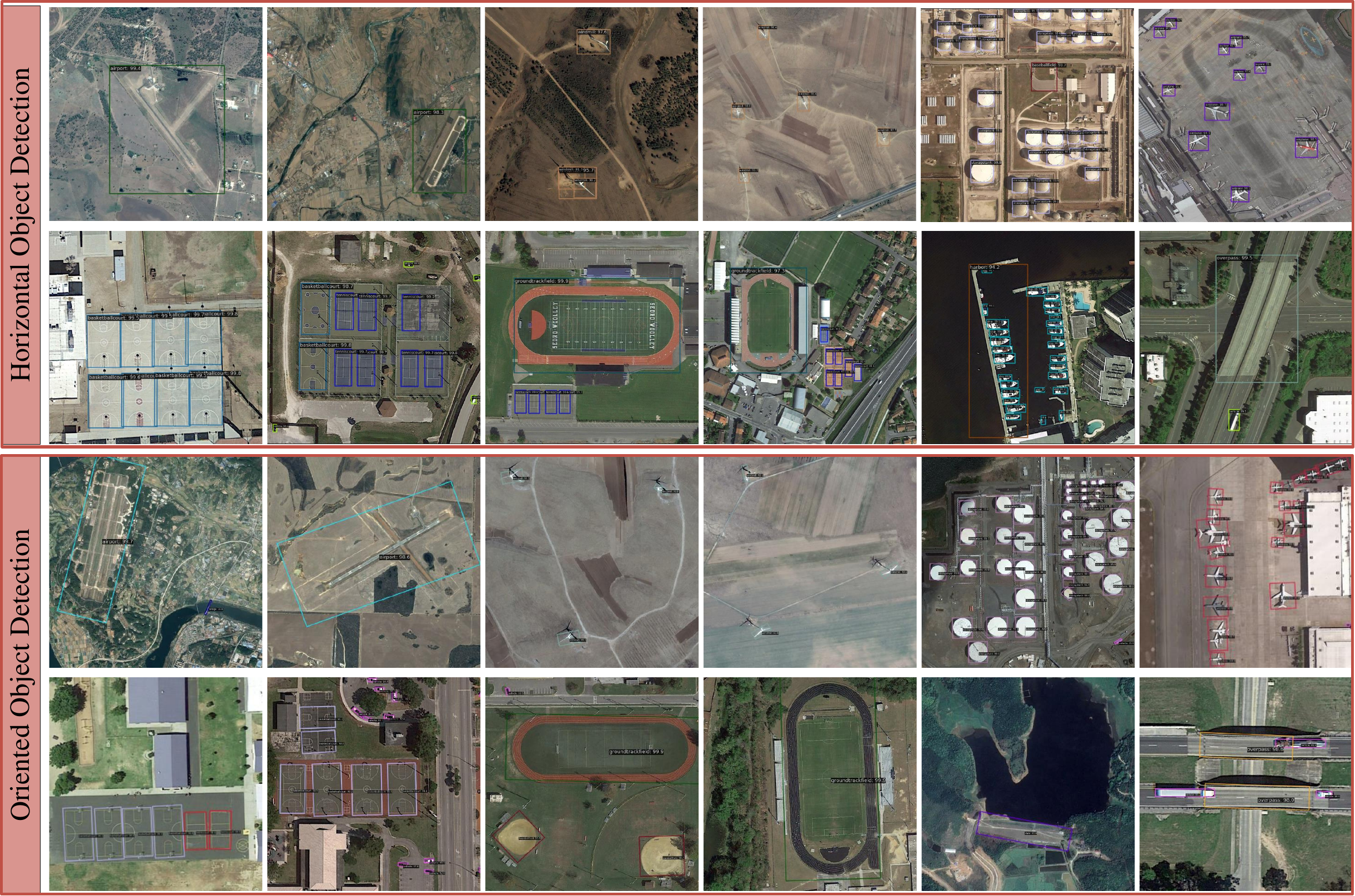}
   \caption{Visualization of aerial open-vocabulary detection inference. Top: Open-vocabulary horizontal object detection results. Bottom: Open-vocabulary oriented object detection results. Novel categories in this figure include airport, basketball court, ground track field, and windmill.}
   \label{fig:inference_vis}
\end{figure*}

\begin{table*}[!ht]
\renewcommand{\arraystretch}{1.2}
\caption{Oriented open-vocabulary detection benchmark on DOTA-v1.0 dataset. The short names for categories are defined as: Plane (PL), Baseball diamond (BD), Ground track field (GTF) and Basketball court (BC).}
\label{tab:com2best_rovd_dota}
% \vspace{-5pt}
\centering
\resizebox{1.\linewidth}{!}{
\begin{tabular*}{1.28\linewidth}{@{}c|cccc|ccccc|cccc@{}}
\toprule
Method & Detector & Backbone & $\mathcal{T}_{\mathrm{novel}}$ & $\mathcal{I}_{\mathrm{cls}}$ & PL & BD & ... & GTF & BC & mAP & mAP$_{\mathrm{base}}$ & mAP$_{\mathrm{novel}}$ & HM \\ \hline
ViLD~\cite{gu2021open} & Faster-RCNN & R50 & \cmark & \xmark & 88.3 & \textbf{64.7} & ... & 5.5 & 0.4 & 55.1 & 63.2 & 2.9 & 5.6 \\
ViLD~\cite{gu2021open} & Oriented-RCNN & R50 & \cmark & \xmark & \textbf{89.2} & {\ul 64.6} & ... & 4.7 & 0.4 & \textbf{59.0} & \textbf{67.7} & 2.5 & 4.8 \\
GLIP~\cite{9879567glip} & GLIP & R50 & \xmark & \cmark & 86.8 & 64.6 & ... & 9.8 & 6.1 & 57.5 & 65.2 & 8.0 & 14.2 \\
GLIP~\cite{9879567glip} & GLIP & Swin-T & \xmark & \cmark & 71.9 & 20.5 & ... & 3.1 & 5.6 & 31.4 & 35.6 & 4.3 & 7.7 \\
GroundingDINO~\cite{liu2023groundingdino} & GroundingDINO & R50 & \xmark & \cmark & 88.4 & 53.2 & ... & 3.0 & 0.7 & 53.1 & 61.0 & 1.9 & 3.6 \\
GroundingDINO~\cite{liu2023groundingdino} & GroundingDINO & Swin-T & \xmark & \cmark & 88.1 & 58.8 & ... & 0.7 & 10.8 & {\ul 58.0} & {\ul 66.0} & 5.8 & 10.6 \\
\rowcolor[HTML]{EFEFEF} 
CastDet (\textit{S}) & Faster-RCNN & R50 & \cmark & \xmark & 88.5 & 42.0 & ... & 30.5 & 25.7 & 52.8 & 56.6 & 28.1 & 37.6 \\
\rowcolor[HTML]{EFEFEF} 
CastDet (\textit{LT}) & Faster-RCNN & R50 & \cmark & \xmark & 88.5 & 51.2 & ... & 27.6 & 30.7 & 56.1 & 60.3 & 29.2 & 39.3 \\
\rowcolor[HTML]{EFEFEF} 
CastDet (\textit{S}) & Oriented-RCNN & R50 & \cmark & \xmark & {\ul 88.9} & 50.3 & ... & {\ul 32.4} & {\ul 33.3} & 55.8 & 59.3 & {\ul 32.8} & {\ul 42.3} \\
\rowcolor[HTML]{EFEFEF} 
CastDet (\textit{LT}) & Oriented-RCNN & R50 & \cmark & \xmark & 88.6 & 57.5 & ... & \textbf{34.2} & \textbf{37.7} & 57.3 & 60.6 & \textbf{36.0} & \textbf{45.1}   \\ \bottomrule
\end{tabular*}
}
\end{table*}

\begin{table*}[!ht]
\renewcommand{\arraystretch}{1.2}
\caption{Oriented open-vocabulary detection benchmark on DIOR dataset. The short names for categories are defined as: Airplane (APL), Airport (APO), Basketball court (BC), Ground track field (GTF) and Windmill (WM).}
\label{tab:com2best_rovd_dior}
% \vspace{-5pt}
\centering
\resizebox{1.\linewidth}{!}{
\begin{tabular*}{1.36\linewidth}{@{}c|cccc|cccccc|cccc@{}}
\toprule
Method & Detector & Backbone & $\mathcal{T}_{\mathrm{novel}}$ & $\mathcal{I}_{\mathrm{cls}}$ & APL & ... & APO & BC & GTF & WM & mAP & mAP$_{\mathrm{base}}$ & mAP$_{\mathrm{novel}}$ & HM \\ \hline
ViLD~\cite{gu2021open} & Faster-RCNN & R50 & \cmark & \xmark & 52.3 & ... & 1.1 & 2.1 & 27.2 & {\ul 6.1} & 37.5 & 44.6 & 9.1 & 15.2 \\
ViLD~\cite{gu2021open} & Oriented-RCNN & R50 & \cmark & \xmark & 53.6 & ... & 0.2 & 2.3 & 16.3 & 9.1 & 39.3 & 47.4 & 7.0 & 12.2 \\
GLIP~\cite{9879567glip} & GLIP & R50 & \xmark & \cmark & 43.3 & ... & 0.1 & 1.6 & 20.2 & 0.0 & 39.4 & 47.9 & 5.5 & 9.8 \\
GLIP~\cite{9879567glip} & GLIP & Swin-T & \xmark & \cmark & 33.3 & ... & 0.1 & 2.6 & 13.8 & 0.0 & 28.7 & 34.8 & 4.1 & 7.4 \\
GroundingDINO~\cite{liu2023groundingdino} & GroundingDINO & R50 & \xmark & \cmark & 43.7 & ... & 0.1 & 0.4 & 10.8 & 0.0 & 39.5 & 48.7 & 2.8 & 5.3 \\
GroundingDINO~\cite{liu2023groundingdino} & GroundingDINO & Swin-T & \xmark & \cmark & 48.0 & ... & 0.9 & 7.3 & 8.4 & 0.1 & {\ul 45.0} & \textbf{55.3} & 4.2 & 7.8 \\
\rowcolor[HTML]{EFEFEF} 
CastDet (\textit{S}) & Faster-RCNN & R50 & \cmark & \xmark & 52.9 & ... & 1.7 & \textbf{51.1} & 26.8 & 1.6 & 40.3 & 45.3 & 20.3 & 28.0 \\
\rowcolor[HTML]{EFEFEF} 
CastDet (\textit{LT}) & Faster-RCNN & R50 & \cmark & \xmark & \textbf{53.9} & ... & 9.1 & 45.4 & 27.2 & 2.8 & 44.3 & 50.1 & 21.1 & 29.7 \\
\rowcolor[HTML]{EFEFEF} 
CastDet (\textit{S}) & Oriented-RCNN & R50 & \cmark & \xmark & 53.2 & ... & \textbf{9.8} & 41.5 & {\ul 27.3} & 9.1 & 42.8 & 48.0 & {\ul 21.9} & {\ul 30.1} \\
\rowcolor[HTML]{EFEFEF} 
CastDet (\textit{LT}) & Oriented-RCNN & R50 & \cmark & \xmark & {\ul 53.8} & ... & {\ul 9.2} & {\ul 49.8} & \textbf{29.3} & \textbf{9.1} & \textbf{45.9} & {\ul 51.3} & \textbf{24.3} & \textbf{33.0}  \\ \bottomrule
\end{tabular*}
}
\end{table*}

\begin{table*}[!ht]
\renewcommand{\arraystretch}{1.2}
\caption{Oriented open-vocabulary detection benchmark on STAR dataset. The short names for categories are defined as: Ship (SH), Airplane (PL), Cooling tower (CT), and Basketball court (BC).}
\label{tab:com2best_rovd_star}
% \vspace{-5pt}
\centering
\resizebox{1.\linewidth}{!}{
\begin{tabular*}{1.3\linewidth}{c|cccc|ccccc|cccc}
\toprule
Method                                    & Detector      & Backbone & $\mathcal{T}_{\mathrm{novel}}$ & $\mathcal{I}_{\mathrm{cls}}$ & SH            & PL            & ... & CT            & BC            & mAP           & mAP$_{\mathrm{base}}$ & mAP$_{\mathrm{novel}}$ & HM            \\ \hline
ViLD~\cite{gu2021open}                    & Faster-RCNN   & R50      & \cmark                         & \xmark                       & 37.1          & \textbf{80.6} & ... & 30.6          & 4.5           & 16.1          & 18.4                  & 4.4                    & 7.1           \\
ViLD~\cite{gu2021open}                    & Oriented-RCNN & R50      & \cmark                         & \xmark                       & \textbf{45.4} & {\ul 80.5}          & ... & 0.0           & 0.9           & 17.1          & 20.5                  & 0.1                    & 0.2           \\
GLIP~\cite{9879567glip}                   & GLIP          & R50      & \xmark                         & \cmark                       & 27.1          & 79.3          & ... & 9.7           & 7.5           & 13.4          & 15.1                  & 4.8                    & 7.3           \\
GLIP~\cite{9879567glip}                   & GLIP          & Swin-T   & \xmark                         & \cmark                       & 0.5           & 21.2          & ... & 0.4           & 0.6           & 1.4           & 1.6                   & 0.2                    & 0.4           \\
GroundingDINO~\cite{liu2023groundingdino} & GroundingDINO & R50      & \xmark                         & \cmark                       & 37.3          & 79.6          & ... & 16.3          & 11.9          & 18.6          & 21.3                  & 5.3                    & 8.5           \\
GroundingDINO~\cite{liu2023groundingdino} & GroundingDINO & Swin-T   & \xmark                         & \cmark                       & 38.1          & 79.2          & ... & 33.5          & 12.5          & 17.8          & 19.9                  & 7.6                    & 11.0          \\ \rowcolor[HTML]{EFEFEF}
CastDet (\textit{S})                      & Faster-RCNN   & R50      & \cmark                         & \xmark                       & 37.3          & 80.3          & ... & 25.6          & 18.8          & 16.4          & 18.4                  & 6.4                    & 9.5           \\ \rowcolor[HTML]{EFEFEF}
CastDet (\textit{LT})                     & Faster-RCNN   & R50      & \cmark                         & \xmark                       & 39.5          & {\ul 80.5}          & ... & \textbf{44.0} & 13.9          & 18.6          & 20.6                  & 8.6                    & 12.1          \\ \rowcolor[HTML]{EFEFEF}
CastDet (\textit{S})                      & Oriented-RCNN & R50      & \cmark                         & \xmark                       & 39.7          & \textbf{80.6} & ... & 40.9          & {\ul 38.6}    & {\ul 20.7}    & {\ul 20.9}            & \textbf{19.4}          & \textbf{20.1} \\ \rowcolor[HTML]{EFEFEF}
CastDet (\textit{LT})                     & Oriented-RCNN & R50      & \cmark                         & \xmark                       & {\ul 44.6}          & {\ul 80.5}          & ... & {\ul 44.3}    & \textbf{44.7} & \textbf{21.3} & \textbf{22.0}         & {\ul 18.1}             & {\ul 19.8}   \\ \bottomrule
\end{tabular*}
}
\end{table*}

\subsubsection{Horizontal Object Detection}

\paragraph{Comparison with CLIPs}
We compare our method with RemoteCLIP~\cite{liu2023remoteclip} under two settings: \rmnum{1}) using ground truth bounding boxes as proposals to measure open-vocabulary classification performance, and \rmnum{2}) employing proposals predicted by RPN to assess overall model performance.
Table \ref{tab:com2clip} shows that CastDet outperforms RemoteCLIP in open-vocabulary classification, achieving an improvement from 37.6\% to 47.3\% mAP. Furthermore, CastDet shows significant performance gains when using RPN proposals, with mAP increasing from 11.6\% to 38.1\%. Additional experiments conducted on COCO~\cite{lin2014microsoft} further validate the improvements achieved by our approach on natural images.

\paragraph{Comparison with SOTAs}
In OVD, auxiliary weak supervision may be employed in various forms: \textbf{1)} leveraging both base and novel categories ($\mathcal{T}_{\mathrm{novel}}$) with unlabeled images for knowledge distillation or pseudo-labeling, such as ViLD~\cite{gu2021open} and OV-DETR~\cite{zang2022open}, which is consistent with our setting; and \textbf{2)} using image classification or image caption datasets  ($\mathcal{I}_{\mathrm{cls}}$) that contain novel concepts, such as Detic~\cite{zhou2022detic}, etc~\cite{Cheng2024YOLOWorld,liu2023groundingdino,9879567glip,wu2023baron}. To support these approaches, we provide the NWPU-RESISC45~\cite{NWPU_RESISC45_7891544} classification dataset. We conduct experiments on DIOR and VisDroneZSD, as shown in Table~\ref{tab:com2best}. Our method outperforms the previous SOTA by 26.8\% mAP$_{\mathrm{novel}}$. Visualization results are presented in Fig.~\ref{fig:inference_vis}.

\paragraph{Comparison with Semi-supervised Methods}

Table~\ref{tab:com_semi} compares CastDet with representative semi-supervised object detection methods, including Instant-Teaching~\cite{zhou2021instant}, Unbiased Teacher~\cite{liu2021unbiased}, and Soft-Teacher~\cite{softteacher_xu2021end}. 
While these methods achieve reasonable recall on base categories, their performance on novel categories completely collapses (mAP$_{\mathrm{novel}}$).
In contrast, CastDet improves both recall and precision, achieving 40.3 mAP$_{\mathrm{novel}}$ and 39.1 HM, while also boosting AR$_{\mathrm{novel}}$ to 69.1. 
These results highlight the clear advantage of CastDet for open-vocabulary detection, especially in discovering novel categories.

\begin{table*}[!tb]
\caption{Comparison with semi-supervised methods on VisDroneZSD.} %\Checkmark
% \vspace{-5pt}
\centering
\resizebox{\linewidth}{!}{
\begin{tabular*}{0.92\linewidth}{c|cccc|cccc}
\toprule
Method                               & AR$_{\mathrm{all}}$ & AR$_{\mathrm{base}}$ & AR$_{\mathrm{novel}}$ & HM            & mAP         & mAP$_{\mathrm{base}}$ & mAP$_{\mathrm{novel}}$ & HM            \\ \hline
Instant-Teaching~\cite{zhou2021instant}                     & 60.0                  & 60.0                   & 61.6                    & 60.5          & 23.2          & 29.0                    & 0.0                      & 0.0           \\
Unbiased Teacher~\cite{liu2021unbiased} & \textbf{64.2}         & \textbf{64.4}          & 63.7                    & 64.0          & 24.0          & 30.1                    & 0.0                      & 0.0           \\
Soft-Teacher~\cite{softteacher_xu2021end}                         & 60.1                  & 63.2          & 47.7                    & 54.4          & 28.7          & 35.9                    & 0.0                      & 0.0           \\
\rowcolor[HTML]{EFEFEF}
CastDet(Ours)                        & 63.6         & 62.2                   & \textbf{69.1}           & \textbf{65.5} & \textbf{38.4} & \textbf{37.9}           & \textbf{40.3}            & \textbf{39.1} \\ 
 \bottomrule
\end{tabular*}
}
% \vspace{-5pt}
  \label{tab:com_semi}
\end{table*}

\subsubsection{Oriented Object Detection}
% Results for DIOR, DOTA.
Given the absence of an oriented open-vocabulary detector, we adhere to the architectures used for horizontal detection and extend them for oriented detection. We compare the proposed method with Oriented ViLD, Oriented GLIP and Oriented GroundingDINO. A benchmark for oriented open-vocabulary detection is established on three oriented datasets: DOTA-v1.0, DIOR, and STAR, with results shown in Table~\ref{tab:com2best_rovd_dota}, Table~\ref{tab:com2best_rovd_dior} and Table~\ref{tab:com2best_rovd_star}. Our method achieves competitive performance, e.g., 36.0 \% mAP$_{\mathrm{novel}}$ on DOTA-v1.0, 
24.3\% mAP$_{\mathrm{novel}}$ on DIOR and 19.4\% mAP$_{\mathrm{novel}}$ on STAR. Fig.~\ref{fig:inference_vis} shows the visualization results.

\subsubsection{Natural Images Open-vocabulary Detection}
\begin{figure}[!h]
  \centering
\includegraphics[width=0.8\linewidth]{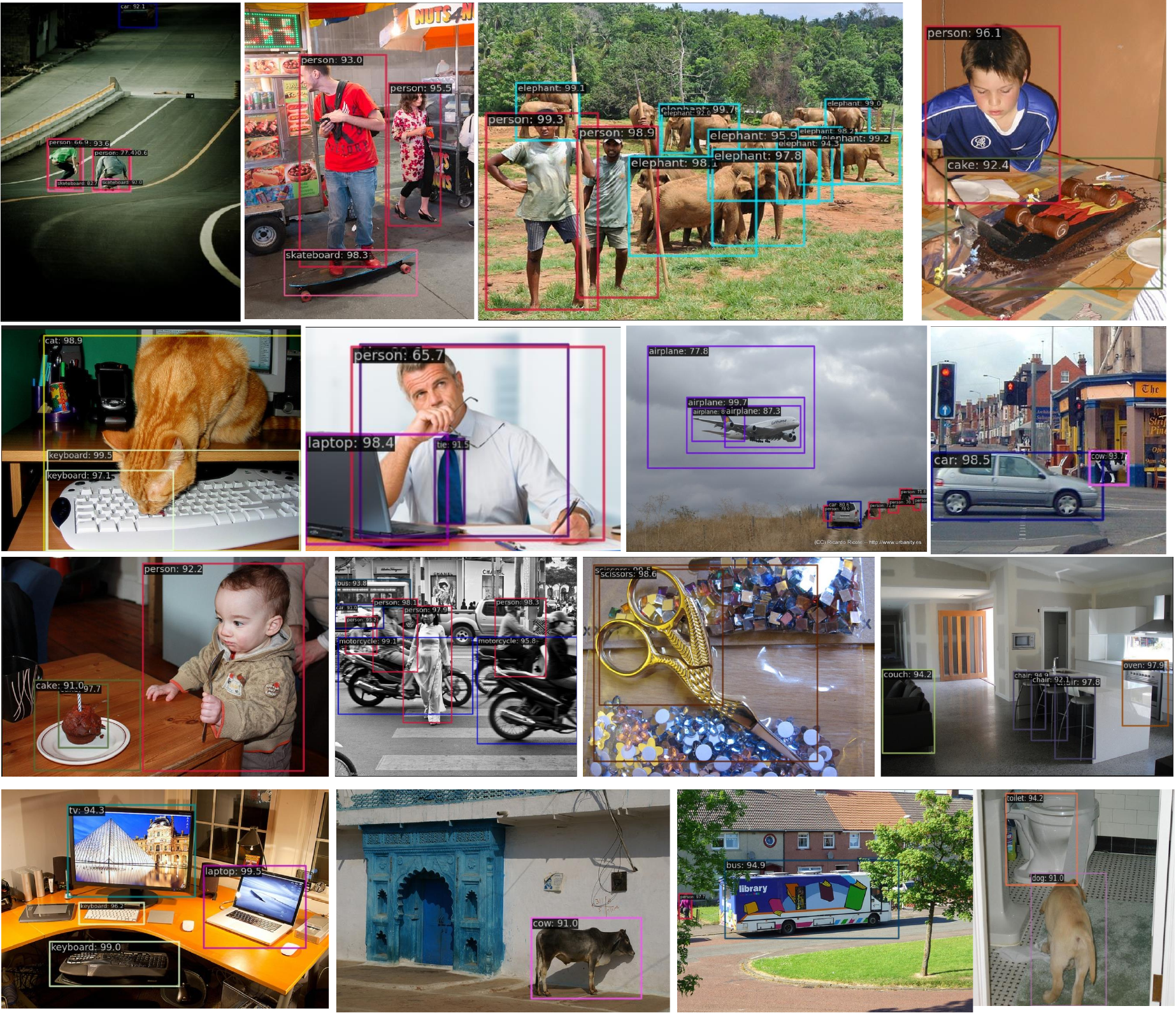}
   \caption{Visualization of open-vocabulary COCO inference. Novel categories in this figure include skateboard, elephant, cake, keyboard, cat, tie, airplane, cow, bus, scissors, couch, and dog.}
   \label{fig:coco_inference_vis}
\end{figure}

To further assess the effectiveness of the proposed method beyond aerial imagery, we extend our evaluations to include the COCO dataset. As shown in Table~\ref{tab:coco_results}, our method achieves impressive results in novel category detection, reaching a 30.3\% mAP$_{\mathrm{novel}}$. The qualitative results are shown in Fig.~\ref{fig:coco_inference_vis}.

\begin{table}[!h]
\centering
\caption{Results on open-vocabulary COCO. $\dagger$: Results quoted from \cite{zareian2021open}. $\ddagger$: The results of our own implementation, under the same experimental setup as ours. $*$: Results quoted from the original paper.}
\label{tab:coco_results}
\begin{tabular}{l|cccc}
\toprule
             & mAP$_{\mathrm{novel}}$ & {\color[HTML]{9B9B9B} mAP$_{\mathrm{base}}$} & {\color[HTML]{9B9B9B} mAP50$_{\mathrm{all}}$} & {\color[HTML]{9B9B9B} HM}   \\ \hline
Mask R-CNN~\cite{maskrcnn_he2017mask}    & -                    & {\color[HTML]{9B9B9B} 51.4}                  & {\color[HTML]{9B9B9B} -}                   & {\color[HTML]{9B9B9B} -} \\
WSDDN~\cite{bilen2016weakly}$\dagger$        & 19.7                   & {\color[HTML]{9B9B9B} 19.6}                  & {\color[HTML]{9B9B9B} 19.6}                   & {\color[HTML]{9B9B9B} 19.6} \\
Cap2Det~\cite{ye2019cap2det}$\dagger$      & 20.3                   & {\color[HTML]{9B9B9B} 20.1}                  & {\color[HTML]{9B9B9B} 20.1}                   & {\color[HTML]{9B9B9B} 20.2} \\
OVR-CNN~\cite{zareian2021open}$^*$      & 22.8                   & {\color[HTML]{9B9B9B} 46.0}                  & {\color[HTML]{9B9B9B} 39.9}                   & {\color[HTML]{9B9B9B} 30.5} \\
Detic~\cite{zhou2022detic}$^*$        & 24.1                   & {\color[HTML]{9B9B9B} 52.0}                  & {\color[HTML]{9B9B9B} 44.7}                   & {\color[HTML]{9B9B9B} 32.9} \\
ViLD~\cite{gu2021open}$\ddagger$ & 12.9                   & {\color[HTML]{9B9B9B} 38.8}                  & {\color[HTML]{9B9B9B} 32.0}                   & {\color[HTML]{9B9B9B} 19.3} \\
PromptDet~\cite{feng2022promptdet}$^*$    & 26.6                   & {\color[HTML]{9B9B9B} 59.1}                  & {\color[HTML]{9B9B9B} 50.6}                   & {\color[HTML]{9B9B9B} 36.7} \\ 
\rowcolor[HTML]{EFEFEF} 
CastDet (Ours)         & \textbf{30.3}          & {\color[HTML]{9B9B9B} 47.4}                  & {\color[HTML]{9B9B9B} 42.9}                   & {\color[HTML]{9B9B9B} 37.0} \\ \bottomrule
\end{tabular}
\end{table}

\section{Discussion and Conclusion}
In this paper, we address open-vocabulary aerial object detection, tackling challenges such as limited annotated data and unique characteristics in aerial scenarios. We present CastDet, a CLIP-activated student-teacher detector that supports both horizontal and oriented detection. By incorporating a flexible student-(multi)teacher self-learning approach, a dynamic label queue, and several box selection strategies, our method effectively bridges the gap between general open-vocabulary detection and the specific needs of aerial imagery.

Beyond its strong performance on two-stage detectors, such as Faster R-CNN and Oriented R-CNN, CastDet also exhibits potential for adaptation to other detection frameworks. Specifically, by adjusting the representation of the confidence score, our method can be transferred to one-stage detectors, where the final confidence score can serve as a substitute for the RPN score. Likewise, for DETR-based detectors, scores such as RJV, BJV, SJV, and AJV can be derived by computing the variance of regression predictions across different transformer layers, providing an effective measure of prediction uncertainty.

Extensive experiments on multiple aerial datasets, including VisDroneZSD, DIOR and DOTA, validate that CastDet achieves significant performance improvements. To the best of our knowledge, this marks the first effort in open-vocabulary aerial detection. We aspire for the proposed techniques and the established benchmark to lay a solid foundation for future research in this area.

% \section{Data Availability}
% The DOTA~\cite{DOTA8578516}, DIOR~\cite{dior_li2020object}, VisDroneZSD~\cite{VisDrone2023}, and NWPU-RESISC45~\cite{NWPU_RESISC45_7891544} datasets used in this study are publicly available aerial datasets. The code for our method and data processing is openly accessible at \url{https://github.com/VisionXLab/CastDet}.

\clearpage

\bibliographystyle{plainnat}
\bibliography{references}

% 如果需要附录，取消下面的注释
% \clearpage
% \beginappendix
% \input{sections/appendix}

\end{document}